\documentclass[conference]{IEEEtran}
\IEEEoverridecommandlockouts
\usepackage{amsmath,amssymb,amsfonts}
\usepackage{graphicx}
\usepackage{url}
\usepackage{tikz}
\usepackage{multirow}
\usepackage{textcomp}
\usepackage{xcolor}
\usepackage{lipsum}
\usepackage{natbib}
\usepackage{xparse}
\usepackage{xspace}
\usepackage{graphicx}
\usepackage{subcaption}
\usepackage{algorithm}
\usepackage{algorithmicx}
\usepackage{algpseudocode}
\usepackage{scalefnt}
\usepackage{array}

\setcitestyle{square}
\newcommand{\ie}{\textit{i.e.,}\xspace}
\newcommand{\eg}{\textit{e.g.,}\xspace}

\def\BibTeX{{\rm B\kern-.05em{\sc i\kern-.025em b}\kern-.08em
    T\kern-.1667em\lower.7ex\hbox{E}\kern-.125emX}}

\newcommand\copyrightnotice{
    \begin{tikzpicture}[remember picture,overlay]
        \node[anchor=south,yshift=10pt] at (current page.south) {\fbox{\parbox{\dimexpr\textwidth-\fboxsep-\fboxrule\relax}{\copyrighttext}}};
    \end{tikzpicture}
}

\newcommand\copyrighttext{
  \footnotesize \textcopyright 2012 IEEE. Personal use of this material is permitted.
  Permission from IEEE must be obtained for all other uses, in any current or future
  media, including reprinting/republishing this material for advertising or promotional
  purposes, creating new collective works, for resale or redistribution to servers or
  lists, or reuse of any copyrighted component of this work in other works.
  DOI: UNDER REVIEW
}

\begin{document}

\title{Multi-level Cellular Automata for FLIM networks}

\author{
\IEEEauthorblockN{Felipe Crispim R. Salvagnini\IEEEauthorrefmark{1}, Jancarlo F. Gomes\IEEEauthorrefmark{1},
Cid A. N. Santos\IEEEauthorrefmark{2}, Silvio Jamil F. Guimarães\IEEEauthorrefmark{3}, \\ Alexandre X. Falcão\IEEEauthorrefmark{1}
}
\IEEEauthorblockA{\IEEEauthorrefmark{1}\textit{University of Campinas}}
\IEEEauthorblockA{\IEEEauthorrefmark{2}\textit{Eldorado Institute}}
\IEEEauthorblockA{\IEEEauthorrefmark{3}\textit{Pontifical Catholic University of Minas Gerais}}
\IEEEauthorblockA{Email: felipe.salvagnini@students.ic.unicamp.br,
jgomes@ic.unicamp.br, 
cid.santos@eldorado.org.br, \\
sjamil@pucminas.br,
afalcao@unicamp.br
}
}
\maketitle
\copyrightnotice

\begin{abstract}

The necessity of abundant annotated data and complex network architectures presents a significant challenge in deep-learning Salient Object Detection (deep SOD) and across the broader deep-learning landscape. This challenge is particularly acute in medical applications in developing countries with limited computational resources. Combining modern and classical techniques offers a path to maintaining competitive performance while enabling practical applications. Feature Learning from Image Markers (FLIM) methodology empowers experts to design convolutional encoders through user-drawn markers, with filters learned directly from these annotations. Recent findings demonstrate that coupling a FLIM encoder with an adaptive decoder creates a flyweight network suitable for SOD, requiring significantly fewer parameters than lightweight models and eliminating the need for backpropagation. Cellular Automata (CA) methods have proven successful in data-scarce scenarios but require proper initialization --- typically through user input, priors, or randomness. We propose a practical intersection of these approaches: using FLIM networks to initialize CA states with expert knowledge without requiring user interaction for each image. By decoding features from each level of a FLIM network, we can initialize multiple CAs simultaneously, creating a multi-level framework. Our method leverages the hierarchical knowledge encoded across different network layers, merging multiple saliency maps into a high-quality final output that functions as a CA ensemble. Benchmarks across two challenging medical datasets demonstrate the competitiveness of our multi-level CA approach compared to established models in the deep SOD literature.

\end{abstract}


\section{Introduction}
\label{sec:intro}

When a specialist analyzes an image, such as a neuroradiologist investigating tumors in patients' Magnetic Resonance Imaging (MRI) data, abnormalities usually stand out and are easily spotted. Likewise, a parasitologist can quickly locate parasite eggs in microscopic images. In some images, even someone without training can identify the abnormalities. Additionally, we naturally perform this task when we observe a scene and the foreground becomes prominent. Salient Object Detection (SOD) methods propose to tackle the same problem, generating a saliency map where the foreground (or abnormality) is brighter than the background.

SOD methods are categorized into bottom-up or top-down approaches \citep{Qin2018}. Bottom-up methods employ low-level features --- local properties such as color, texture, and descriptors --- and statistics to identify foreground regions \citep{botton_up}. Conversely, top-down techniques utilize high-level features extracted through supervised methods, which require annotated data \citep{basnet, u2net, sod_deep}. Some works also categorize those approaches into traditional manual feature learning and deep-learning-based methods \citep{fuse_sod}.

Despite the recent success of deep-learning approaches applied to SOD, as models get even more sophisticated, they face two-main challenges under real-world applications: the demand for high-cost computational resources (GPUs) and the requirement of many annotated data (even when pre-trained). The literature proposes to solve the former through \textit{Lighweight Models}, where the complexity of the models is reduced by the model's design or compression \citep{review_lightweight_DCNN_2024}. Although they speed up inferences and reduce model size, lightweight models still have around 1 million to 5 million parameters and require a large amount of annotated data.

The FLIM methodology tackles deep-learning challenges by enabling us to design flyweight convolutional networks (usually 2 to 5 layers) under expert control \citep{joao2023flyweight}. Flyweight is a term that describes networks even lighter than lightweight models, with around 500k parameters. Inserting a user in the interactive pipeline enables: (1) Selection of representative images (usually less than 10) \citep{abrantes_sibgrapi_2024}; (2) Design of a convolutional encoder from weak annotations inserted by a user (\eg scribbles or disks) \citep{crispim_sibgrapi_2024, gilson_sibgrapi_2024}; (3) Get the saliency map by adaptively decoding features extracted by the encoder \citep{gilson_sibgrapi_2024}. FLIM is a recently proposed method, and each of those three steps is under investigation.

FLIM networks are suitable for data-scarcity scenarios where the entire network is backpropagation-free. Hence, there is no requirement for densely annotated ground truth, only weak annotations on a few images. Such an approach is not data-hungry and employs learned high-level features to generate saliency maps. The saliency map generated by FLIM is particularly suitable to initialize a bottom-up method. The latter behaves as a post-processing operation, where we take advantage of low-level and high-level features.

Cellular Automata (CA) methods present an excellent candidate for integration with FLIM networks as a post-processing mechanism, where FLIM networks can significantly enhance the initialization phase. CA is commonly applied to data-scarce scenarios (\eg medical images), where the CA is executed for each input. CA also enables SOD methods using both low-level \citep{grow_cut, tumor_cut} and high-level features \citep{Qin2018, Quin2015}. Nevertheless, CA methods require proper initialization. Common strategies are: initialize states from user inputs like a line drawn on image regions \citep{grow_cut, tumor_cut, improved_ca}; random initialization \citep{unsup_ca}; priors (as background or contrast priors) \citep{Qin2018}. As far as we are concerned, no work takes advantage of high-level features during initialization; only user-based methods employ the expert's knowledge but require the user for every inference image. Therefore, investigating FLIM for CA initialization is of particular interest.

The meeting of FLIM networks with CA enables a novel initialization technique, characterized by three phases: (1) Design of a FLIM Encoder, which together with an adaptive decoder yields a FLIM network; (2) at inference time initialize CA using a FLIM-Network saliency; (3) evolve the CA to generate a better saliency map. \textbf{The key aspect is, unlike other user-interaction-based SOD methods, user interaction is not required at inference time}.

Our previous research investigated the initialization of CA using FLIM networks \citep{crispim_sibgrapi_2024}. Specifically, with an $L$-layer FLIM encoder, we decoded feature maps from the last convolutional layer into a saliency map to initialize the CA. As established in \citep{crispim_sibgrapi_2024} and reiterated here, FLIM encoders detect descriptive and discriminative patterns beginning from the first layer. The initial layers activate strongly in edge regions, providing sharp features but generating numerous false positive activations. In contrast, deeper layers produce fewer or no false positives, though they yield blurrier activations in edge regions. These complementary characteristics suggest the value of decoding saliency maps from each FLIM-Encoder level to initialize multiple CA simultaneously. This multi-level initialization approach enables exploring a broader initialization space, combining the precision of early layers with the reliability of deeper representations, better exploring the edge and internal regions of the salient objects. Moreover, this novel method, called multi-level CA, opens room for investigation on combining the saliency of multiple CAs, which can be seen as an ensemble of CA models.

In improving our previous study, this work investigates the applicability and generalization of our multi-level CA towards two challenging datasets: brain tumors (glioblastomas) and parasite eggs (Schistosoma Mansoni) detection. It evaluates the proposed method towards gray-scale MRI images ---  with high heterogeneity from multiple institution acquisitions \citep{brats2021} --- and RGB microscopy images \citep{SuzukiTBE2013}. Our main contributions are:

\begin{enumerate}
    \item We demonstrate significant performance enhancements by initializing multi-level CA with saliencies from each level of the FLIM network. Measurable improvements are observed across each hierarchical layer;
    
    \item We extend compassion to SOD literature to include lightweight models to address the criticality of deploying salient object detection in low-cost devices;
    
    \item We propose a simple yet elegant fusion mechanism --- only three convolutional filters --- that effectively integrates multi-level saliency maps into a final saliency. Our experimental results show that our method achieves comparable performance or even surpasses specialized lightweight models and state-of-the-art deep learning SOD architectures (pre-trained on thousands to millions of images).
\end{enumerate}

This work is organized as follows: Section \ref{sec:2} provides a comprehensive review of SOD literature, covering both classical bottom-up models and modern top-down deep learning approaches. Additionally, it introduces the FLIM methodology and explores CA applications in both SOD and semantic segmentation problems, establishing the foundation necessary for understanding our proposed method. Section \ref{sec:3} presents our multi-level CA method in detail, examining the integration of FLIM networks into hierarchical CA initialization, the CA evolution process, and the ensemble technique that combines multiple CAs to yield enhanced saliency maps. Section \ref{sec:4} describes our experimental setup, guiding the reader through creating a  multi-level CA. At the same time, we also discuss the methodology for benchmarking our approach against state-of-the-art methods. Section \ref{sec:5} thoroughly discusses our results across each stage of the proposed method. Finally, Section \ref{sec:6} concludes with our findings, acknowledges the limitations of our approach, and outlines directions for future research.

\section{Related works}
\label{sec:2}

This section reviews the bottom-up and deep-learning SOD methods and builds the intuition behind FLIM methodology and Cellular Automata, setting the foundation for our approach.

\subsection{Bottom-up methods}

Bottom-up methods are rooted in cognitive psychology and human perception to predict which areas of an image capture people's attention \citep{itti_base_sod}. This process typically consists of two stages: first, identifying the region that stands out, and then accurately segmenting that region by fitting a saliency map to the object's interior and edges \citep{sod_review_all, sod_review_deep, sod_deep}.

These classical methods operate directly on low-level features, such as image intensity, color, contrast, and gradients \citep{itti_base_sod, wang2024salient}. Hence, they do not need to learn semantic information, alleviating the requirement for abundant training data. Intrinsic features (derived from the image itself) and/or extrinsic features (\eg statistics from images or user interaction) are employed, where computational heuristics or classical supervision techniques yield the final saliency map.

\cite{itti_base_sod} drew attention to the topic by taking inspiration from the efficiency of primates' visual mechanism for complex scene understanding. Their seminal work operates over colors, intensity, and orientations at multiple scales (\ie gaussian pyramids) to generate a salience map. The first SOD models were based on local contrast — the difference between a pixel, or a patch, and its neighbors — to compute salient regions \citep{bottom_up_4, bottom_up_5, bottom_up_2, bottom_up_1, bottom_up_3}.

For example, \cite{bottom_up_4} employs a perceive field (same size as the input image), where each pixel is a perception unit operating on a neighborhood, measuring the contrast as the color distance in LUV space. This local contrast saliency map is then processed by a region-growing process (fuzzy growing), which splits the pixels into two classes: attended or unattended areas.  \cite{bottom_up_2} further expands by applying Conditional Randon Field (CRF) learning to combine local, regional, and global features: multi-scale contrast, center-surround histogram, and color spatial distribution. \cite{bottom_up_1} adds and normalizes multi-scale saliency map, where a thresholded image generates the salient object map. Later, \cite{bottom_up_3} improves the use of image features by applying edge detection, threshold decomposition, and distance transformation \citep{dist_trans}. Remarkably, most of the preliminary works on the field employed intrinsic features (input image only) and center-pixel neighborhoods despite being computed at multiple scales.

Other works then explored whole image regions, such as the work of \cite{bottom_up_5}, where a hybrid approach first extracts an intermediary saliency using contrast and later improves by averaging the saliency value of the image's regions, where the mean shift image segmentation algorithm extracts the regions. Another example is the work of \cite{bottom_up_6}, which efficiently computes pixel-level global contrast (using a global histogram) and later improves the intermediary results with a region-level analysis and spatial weighting. This work extracts regions using Felzenszwalb and Huttenlocher's segmentation method \citep{felzenszwalb2004efficient}. Concurrently, as more images became available, researchers began incorporating extrinsic features, enabling salient object detection methods to leverage similar images as references to generate saliency maps \citep{bottom_up_7}.

A deeper review of SOD literature will also reveal that some assumptions improve the overall results. As is the case of \textit{background prior} --- assuming that pixels/blocks/regions at the image frame are the foreground, or conversely the \textit{center prior}. However, they do not stand for all categories of images (\eg parasites in microscopy images can appear near the image's frame). For readers seeking a more comprehensive review of the techniques employed by bottom-up methods, we strongly recommend the survey conducted by \cite{sod_review_all}.

Nevertheless, despite the vast amount of bottom-up methods, the literature has consistently struggled with a critical weakness: poor generalization. As most approaches rely on low-level and hand-crafted features, applying them to different domains was highly challenging, particularly in complex scenes. The emergence of CNNs promoted a significant shift in the field, as researchers began leveraging deep learning ability to automatically extract and integrate low-level and high-level semantic features, producing more accurate and robust saliency maps across diverse visual contexts.

\subsection{Deep Learning \& Lightweight models}

Categorized as top-down approaches, deep-learning-based SOD methods (\textit{deep SOD}) focus on learning high-level semantic features, commonly through supervision. Deep SOD employs an encoder that learns both low-level (\eg edges, corners) and high-level features (\eg complex textures, shapes) and, subsequently, a decoder that combines extracted features into a saliency object map --- the pixel-wise prediction of the most visually important objects in the scene --- with the size of the input image \citep{sod_deep}.

Since 2015, when deep SOD was introduced, the majority of works in the field apply Deep Neural Networks (DNNs) to explore local and global features \citep{first_deep_sod, second_deep_sod, third_deep_sod}. Classical methods in the past had struggled in both tasks, where methods based on local features are usually better at detecting salient objects' edges than their interior; conversely, global methods do not perform well when the object's textures are similar to the background.

Building this evolutionary shift in SOD methodology, \cite{first_deep_sod} proposed a saliency detection approach called LEGS (Local Estimation and Global Search) utilizing two deep-neural networks (DNNs), a local (DNN-L) and a global one (DNN-G). DNN-L focuses on local features, classifying a central pixel (of a $51 \times 51$ patch) as salient/non-salient. DNN-L is applied to the input image through a sliding window at inference and later refined through geodesic object proposal. Globally, a feature vector describes each region's global contrast, geometric information, and local saliency measurements (from DNN-L). DNN-G gets the global feature vector of each region as input and outputs the region's salient value, which is further sorted and yields the final saliency map through a weighted sum of the top K candidate masks. DNN-L comprises convolutional, max-pooling, and fully connected layers, while DNN-G is built only with fully connected layers.

In addition to exploring local and global features, the work of \cite{second_deep_sod} proposes a multiscale approach using CNN features. Employing a neural network architecture with fully connected layers on top of CNNs, they extract features at three different spatial scales, generating saliency maps at different scales, which are then linearly combined. A key innovation was leveraging convolutional weights pretrained on the ImageNet dataset, enabling rich semantic understanding for visual saliency detection. Similarly, \cite{third_deep_sod} also employed a single network to derive a saliency map, but through a dual branch architecture, a local-context branch and a global-context branch, which outputs are employed to classify a super-pixel (the super-pixel's center is the network's input) as salient/non-salient. They also carried out a more in-depth analysis of using pre-trained models.

The onset of larger datasets also played a central role in deep-learning adoption, such as  MSRA-5000 ~\citep{msra_5k}, SOD ~\citep{sod_dataset}, ECCSD ~\citep{eccsd}, PASCAL-S ~\citep{pascal_s}, MSRA10K ~\citep{msra_10k}, and DUTS ~\citep{duts}.

The seminal deep-SOD's works based on multi-layer perceptrons (\ie fully connected layers) took adantage of image subunits and object proposals \citep{first_deep_sod}. Image subunits are super-pixels and patches, while between examples of object proposals, we could cite regions, as the work of \cite{first_deep_sod}, and bounding-boxes. Those subunits are then scored as salient/non-salient. Despite outperforming bottom-up approaches, these methods cannot fully explore spatial information and are time-consuming (\ie running for all image subunits at inference time).

To tackle the aforementioned problems of representation learning and inference time, the literature employed fully convolutional networks (FCN) \citep{dhsnet}. FCNs enable a single model capable of learning multi-level features simultaneously, which better explores spatial information. Furthermore, FCNs operate over the whole input image, requiring a single feed-forward process, significantly speeding up inferences. The seminal work of \cite{dhsnet} proposed the DHSNet (Deep Hierarchical Saliency Network), which is composed of a two-step architecture. First, they coarsely detect salient objects from a global perspective (VGG Convolutional encoder), and then, through recurrent convolutions, they refine the details of the saliency map in a step-by-step fashion.

Recent SOD models further explore multi-scale feature extractions to improve results and also loss functions to better guide the representational learning phase. Great examples are the works of \cite{basnet, u2net}, which proposes the BasNet and U$\mathbf{^2}$-Net architectures. BasNet is a deeply supervised encoder-decoder (\ie it employs deep supervision, computing loss, and backpropagating errors for each decoder level) ~\citep{basnet}. Its weights are learned through a hybrid boundary-aware loss to learn at pixel, patches, and map levels (\ie BCE, SSIM, and IoU, respectively). Extending this approach, U$\mathbf{^2}$-Net takes advantage of nested U-shaped architectures to improve high-level feature extraction at multiple resolutions (intra-stage multi-scale features) ~\citep{u2net}.

Still, there is an ongoing trend to enhance model architecture for better extraction of features at multiple scales. A good example is the work of \cite{Huang2023}, which proposes the Multilevel Diverse Feature Aggregation Network (MDFNet), composed of different modules to capture saliency information at multiple directions and channels, learn correlation of different salient objects, improve extraction of boundary information. The proposed method combines multilevel features and detects suitable feature spaces and channels for better saliency maps. Although this trend yields models suitable for complex scenarios, the downside is higher complexities and, consequently, high inference time.

The problem of high inference time is worst in scenarios where computing resources are limited, such as our problem of detecting parasite eggs in a tropical country (\eg Brazil), where laboratories usually do not have computers with GPUs. To address this issue, one could refer to literature on lightweight models concerned with lowering hardware requirements and speeding up inferences with competitive performance \citep{review_lightweight_DCNN_2024}. The first approach to generate a lightweight model focuses on architecture design, where fewer model parameters and computational complexity are possible through convolution paradigms (\eg dilated separable convolutions \citep{liu2021samnet} and depth-wise convolutions \citep{sandler2018mobilenetv2}) or AutoML \citep{Chu2020MobileAutoML}. On the other hand, the second approach employs model compression by optimizing larger models through model pruning, quantization, knowledge distillation, or low-rank decomposition \citep{Phan2020LR, Chariton2022KD, Huang2024}. Despite those efforts, lightweight models still have around 1M to 5M parameters and require considerable training data. Furthermore, under constrained scenarios, high inference times are still prohibitive (\eg running for a slicing window of a microscopy slide).

To summarize, key questions remain for deep SOD models, mainly in scenarios where we do not have large annotated datasets and inference time is mandatory. Related to datasets, when few annotated data are available, how can we efficiently train a network? Should we consider training from scratch, or would transfer learning be a more practical approach? Regarding the model architecture, how can we discover the lightest network that can solve our problem? How many convolutional layers are required? And the correct number of filters? Those questions permeate not only the deep SOD field but deep learning in general.

\subsection{Feature learning from Image Markers}
\label{sec:flim_rw}

\begin{figure*}[!htpb]
  \centering
  \includegraphics[width=\textwidth]{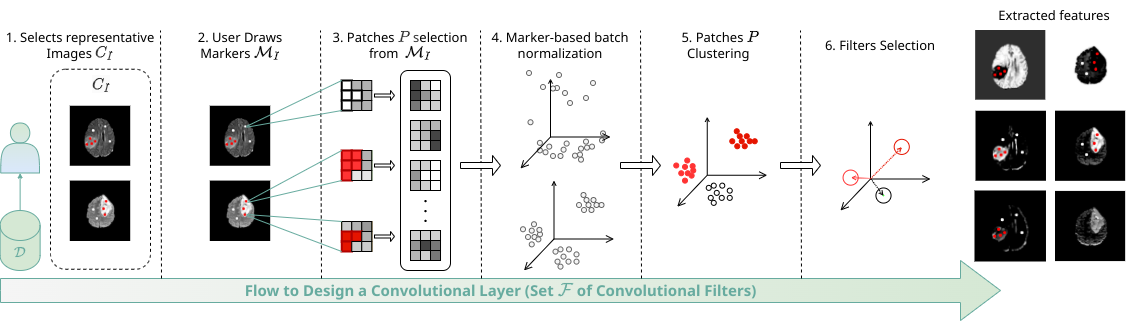}
  \caption{Multi-Level Cellular Automata for FLIM networks: (1) Users iteratively select training images and define FLIM-Encoder architecture. (2) During inference, images enter the encoder, and features decode into intermediary saliency maps, initializing CA cells. (3) Multi-level saliency maps evolve from this initialization, and (4) merge into a final saliency.}
  \label{fig:flim_methodology}
\end{figure*}

Over recent years, works employing the FLIM methodology have consistently demonstrated comparable and even superior results under a data scarcity regime ~\citep{joao2023flyweight, flim_1, flim_2}. Such scenarios are particularly challenging to deep-learning models. Deep SOD models commonly overfit and do not generalize to unseen data when learning with little data. For instance, in \cite{sod_review_deep}, they argue that SOD methods struggle when applied to unseen datasets, even if trained with thousands of data. Hence, at least fine-tuning deep SOD methods towards the target task is a must, but learning a different domain is still a problem.

The FLIM methodology (Figure \ref{fig:flim_methodology}) is a recently proposed approach \citep{flim_2}, which is under active investigation. It integrates user knowledge into the design of a feature extractor (\ie a convolutional encoder). User knowledge is exploited in different methodology phases, encompassing the selection of training images, identification of discriminative regions for filter estimation, and filter evaluation and selection. The methodology allows the design of a CNN encoder layer by layer. The user can create the encoder architecture by selecting representative images and drawing markers on descriptive and discriminative image regions. Conventional filters are then directly estimated from marked regions. Each FLIM encoder holds operations like marker-based normalization, the convolution with a filter bank,  the activation function (\eg, ReLU), and pooling (\eg, max-pooling or averaging pooling). All operations, except marker-based normalization, are well-known in the literature. However, the below definition and geometrical interpretation in light of the FLIM methodology provide vital insights.

Let $\textbf{I}=(D_I, \vec{I}) \in \mathcal{D}$ be an $m$-channel image from dataset $\mathcal{D}$, where $D_I\subset Z^2$ is the image domain and $\vec{I}(p)\in \mathbb{R}^m$ assigns feature values to pixel $p\in D_I$. A patch vector $\vec{P}_p \in \mathbb{R}^{k \times k \times m}$ vectorizes features in a $k \times k \times m$ square region centered at $p$. The same holds for grayscale ($m=1$), color ($m=3$), and feature maps $(m \geq 1)$ from filter bank convolutions. It is important to understand that in Figure \ref{fig:flim_methodology} we represented the FLIM's filter estimation and extracted features given an image as input (first convolutional layer). However, the input features from past layers and markers are projected into the feature space for deeper layers, so filters are learned directly from features.

The selection of training images is possible through multiple approaches: (1) random selection of a fixed number of images, (2) comprehensive dataset inspection to identify representative images --- as demonstrated in our previous research \citep{crispim_sibgrapi_2024} ---, or (3) iterative selection using validation metrics to build the training set progressively. We implemented the iterative approach in this work, although recent studies have demonstrated the effectiveness of supervised and unsupervised techniques for training image selection \citep{abrantes_embc, abrantes_sibgrapi_2024}. This step yelds a subset $C_I \subseteq \mathcal{D}$ (usually, $|C_I| << |\mathcal{D}|$) of a few representative images.

Given $C_I$, users mark discriminative regions (disks, scribbles) on each image $\textbf{I} \in C_I$ while specifying architectural parameters—including filter counts per marker and image, and the target kernel bank size. This process extracts overlapping patches from marked regions, creating a set $\mathcal{M}_I$ of patch vectors $\vec{P}_p$ for pixels $p$ marked in image $\textbf{I}$. These user interactions, required only for training images, demand domain knowledge of both $\mathcal{D}$ and the application context.

With $\mathcal{M} = \bigcup_{\textbf{I} \in C_I} \mathcal{M}_I$, patch vectors in $\mathcal{M}$ are normalized by z-score, and these normalization parameters are applied to all patch vectors from any image $\textbf{I} \in \mathcal{D}$. This \emph{marker-based image normalization} centralizes the dataset $\mathcal{M}$ and corrects distortions along the main axes of $\mathbb{R}^{k\times k \times m}$.

For each marker, the user specifies $n$ filters, and k-means clustering of normalized patch vectors produces $n$ clusters whose centers define kernel weight vectors $\vec{K}_i \in \mathbb{R}^{k \times k \times m}$. These kernels compose the filter bank, which may require further reduction (via k-means or PCA) if the total filter count exceeds target layer specifications or the number of filters by image. Convolving a normalized image $\textbf{I}$ with kernel $\vec{K}_i$ produces an image $\textbf{J}$ where $J_i(p) = \langle\vec{P}_p, \vec{K}i\rangle$. With a filter bank $\{\vec{K}_i\}_{i=1}^{n\times M}$ from $M$ markers and $n$ filters per marker, $\textbf{J}=(D_J,\vec{J})$ becomes a feature map with $n\times M$ channels and $\vec{J}(p)=(J_1(p),J_2(p),\ldots,J_{n\times M}(p)) \in \mathbb{R}^{n \times M}$.

Geometrically, each kernel $\vec{K}_i$ represents a vector normal to a hyperplane through the origin of $\mathbb{R}^{k \times k \times m}$. The patch vector $\vec{P}_p$ exists as a point in this space, with $J_i(p)$ measuring the signed distance from this point to the hyperplane—positive or negative depending on which side the point falls. ReLU operations remove points on the negative side, while max-pooling consolidates nearby activations. Enforcing unit norm $\|\vec{K}_i\| = 1$ across all kernels prevents $J_i(p)$ amplification by kernel magnitude. The marker-based normalization eliminates bias by centering clusters around the origin of $\mathbb{R}^{k \times k \times m}$.

The operations described so far—patch extraction, marker-based normalization, and subsequent clustering for filter estimation from cluster centers—will repeat until the whole encoder architecture is designed. Layers other than layer one will use feature maps of the previous layer (extracted from $\mathcal{C}_I$ images) and project markers into the features' domain. At this point, we have a FLIM encoder suitable for extracting feature maps for any image in $\mathcal{D}$.

\begin{figure}[!ht]
    \centering
    \begin{subfigure}[b]{0.117\textwidth}
        \centering
        \includegraphics[width=\textwidth]{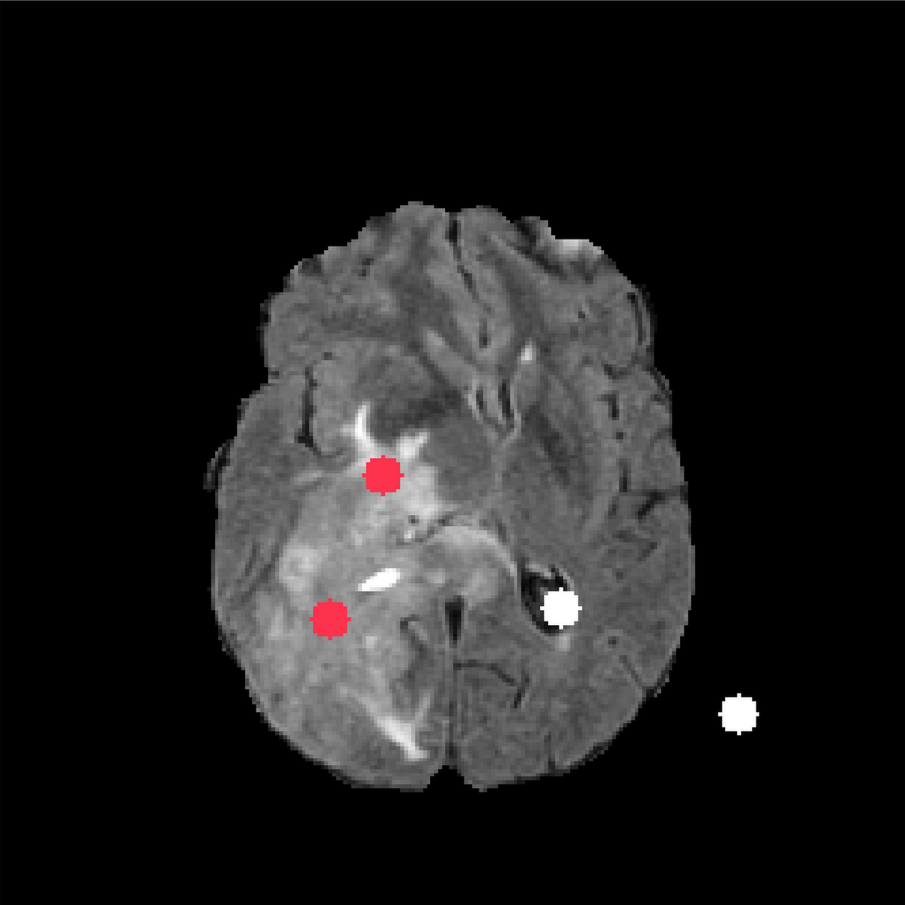}
        \caption{}
        \label{fig:sample_decoder_1}
    \end{subfigure}
    \begin{subfigure}[b]{0.117\textwidth}
        \centering
        \includegraphics[width=\textwidth]{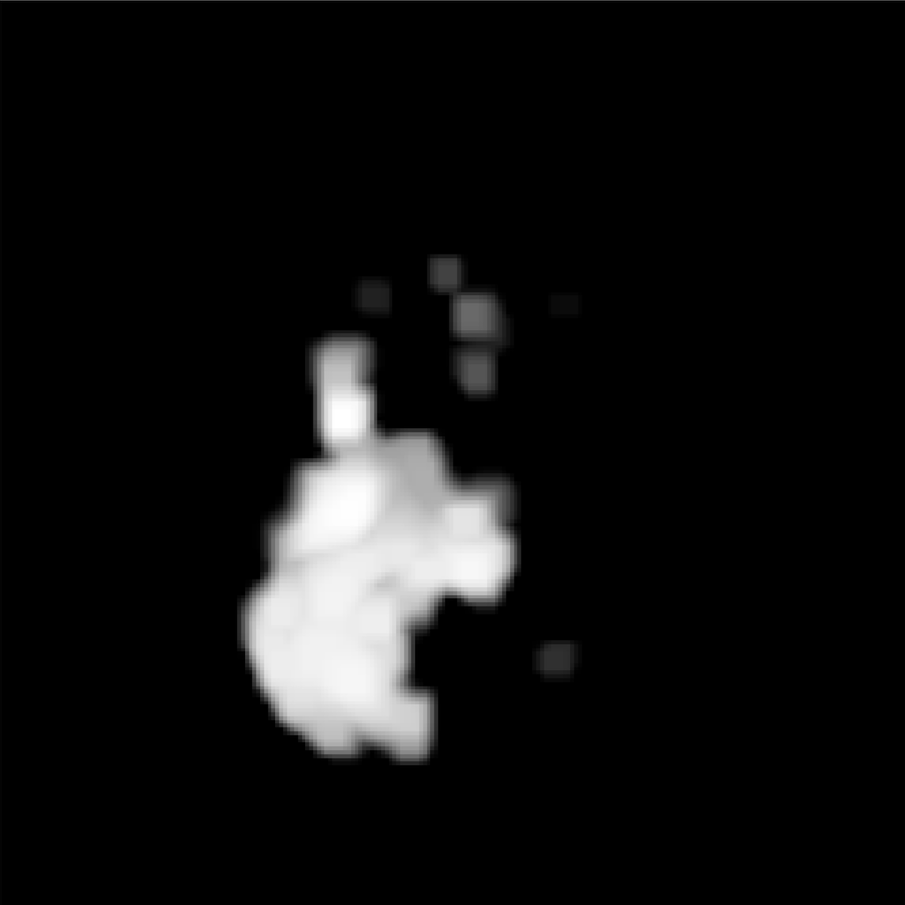}
        \caption{}
        \label{fig:sample_decoder_2}
    \end{subfigure}
    \begin{subfigure}[b]{0.117\textwidth}
        \centering
        \includegraphics[width=\textwidth]{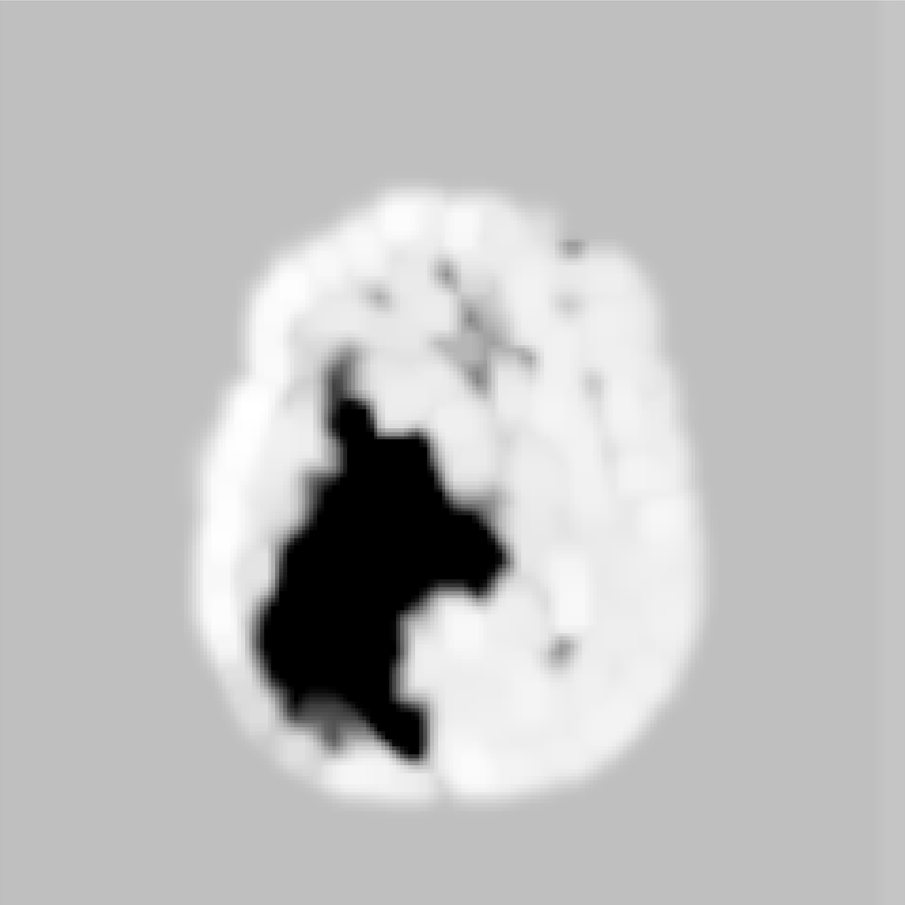}
        \caption{}
        \label{fig:sample_decoder_3}
    \end{subfigure}
    \begin{subfigure}[b]{0.117\textwidth}
        \centering
        \includegraphics[width=\textwidth]{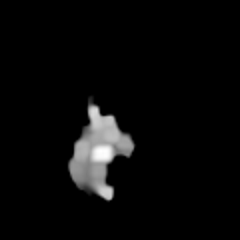}
        \caption{}
        \label{fig:sample_decoder_4}
    \end{subfigure}
    \caption{Adaptive decoder for FLIM-Encoder layer 3 (same across all $l$ layers): (a) input with foreground/background markers; (b) channel $i$ with $w_i = 1$ (foreground); (c) channel $j$ with $w_j = -1$ (background); (d) resulting saliency map $\mathcal{S}_3$.}
    \label{fig:sample_decoder}
\end{figure}

A key property of FLIM encoders highlighted in \cite{joao2023flyweight} involves foreground-background differentiation. When markers are placed in distinct regions (Figure \ref{fig:sample_decoder_1}), the resulting feature maps show channel-specific activation patterns for either foreground or background elements (Figures \ref{fig:sample_decoder_2}-\ref{fig:sample_decoder_3}). For each layer $l$ in $L$ (total encoder layers), saliency maps $\mathcal{S}_l$ can be generated by applying a weighted channel averaging followed by activation function $\phi$ (Figure \ref{fig:sample_decoder_4}). Optimal saliency maps with enhanced object visibility result from decoders assigning positive weights to foreground channels and negative weights to background channels, eliminating false positives from foreground maps. Since channel weight polarity varies with input images, adaptive decoders become necessary. The work of \cite{joao2023flyweight} presents dataset-specific adaptive decoders, but we will follow with a modified version from \cite{Joao2023BuildingFFVisApp}, employed in our previous work \citep{crispim_sibgrapi_2024}, that aligns with our requirements for saliency map $\mathcal{S}_l$ estimation (Equation \ref{eq:decoder}), where \begin{equation}\label{eq:decoder}
\mathcal{S}_l(p) = \phi(\langle \vec{J}_l(p), \vec{w} \rangle ),
\end{equation} where $\vec{w} = (w_1, w_2, \ldots, w_{n \times M})$, $\vec{J}_l(p)$ is the feature vector at position $p$ for layer $l$, and $\phi$ is a given activation function (\eg ReLU). Each feature map channel weight is estimated as $w_i \in \{-1, 0, 1\}$, by \begin{equation}\label{eq:w_adaptive}
w_i = \begin{cases}
    +1,       & \text{if} \quad \mu_{J_i} \leq \mathcal{T} - \sigma^2 \mbox{ and } a_i < A_1,\\
    -1,      & \text{if}  \quad \mu_{J_i} \geq \mathcal{T} + \sigma^2 \mbox{ and } a_i > A_2,\\
    0,       & \text{otherwise}.
\end{cases}
\end{equation}

At this point, we have a FLIM network trained without backpropagation and consequently without requiring many pixel-wisely annotated ground truths. We can now run it in inference for training, validation, and testing images and extract saliency maps at any encoder level (\ie layer).

\subsection{Cellular Automata}

% Very interesting link on CA for further reading (Curiosities): https://seop.illc.uva.nl/entries/cellular-automata/

First introduced in 1951 by John Von Neumann \citep{origin_ca}, Cellular Automata (CA) --- or Cellular Automaton in singular --- refers to a discrete evolving model capable of mapping complex behaviors through relatively simple evolution rules. At its core, a CA consists of a regular arrangement of cells, typically organized in a lattice structure. Common examples include grids of cells (analogous to pixels) for 2D images or cubes of cells (similar to voxels) for 3D images. Nevertheless, cells can have formats other than squares and cubes, as complex surfaces can be represented by irregular tessellations such as Voronoi diagrams or super-pixels \citep{Quin2015}. This adaptability, combined with the ability to model local iterations, which produce global emergent patterns, is suitable for SOD, where identifying visually distinct regions requires both local contrast analysis and global contextual understanding. Hence, by combining CA with FLIM networks, we create a hybrid approach suitable to tackle SOD challenges, where a FLIM network (FLIM convolutional encoder + adaptive decoder) initializes the CA. Thus, this approach leverages both the hierarchical feature learning capabilities of a FLIM network and the iterative, rule-based refinement of a CA.

\begin{figure}[!ht]
  \centering 
  \includegraphics[width=0.48\textwidth]{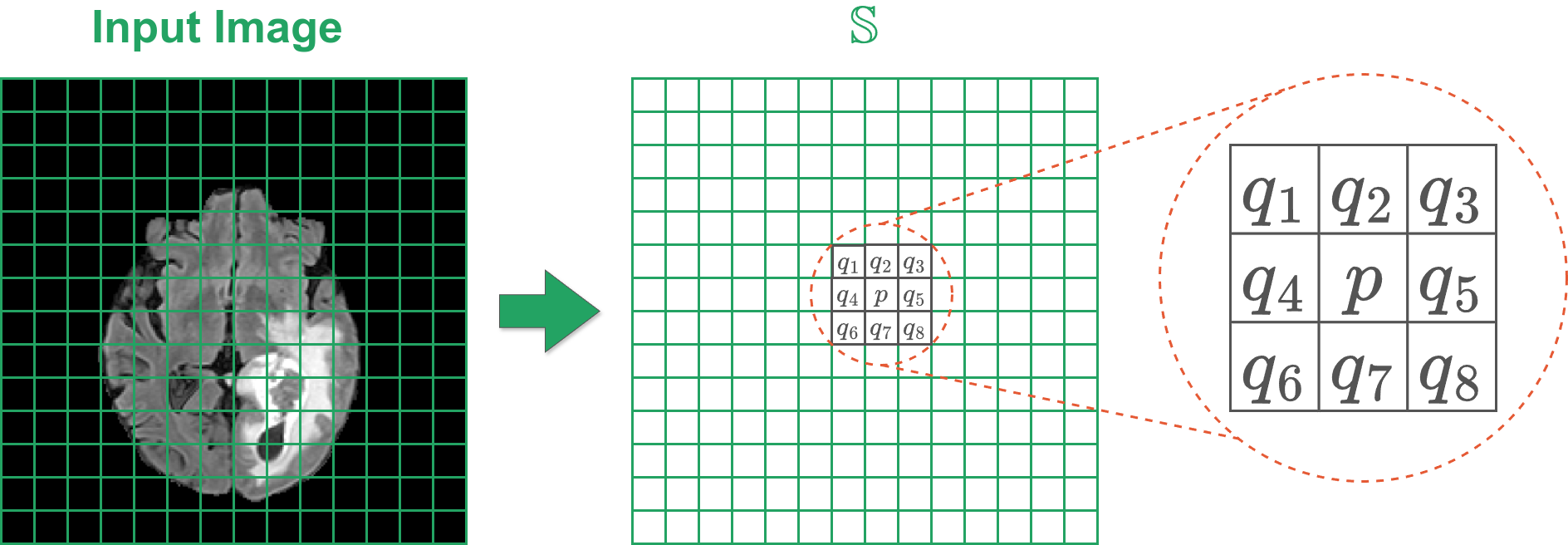}
  \caption{Brain image as a CA's lattice of cells, showing individual cells and their neighboring relationships.}
  \label{fig:ca_sample}
\end{figure}

When operating over images, the CA requires a set $D_I$, representing the image domain containing all cells (lattice of cells). Accordingly, we define the CA as a triple $(\mathbb{S}, N, \delta)$, which operates over the lattice of cells, dictating changes over time given each cell's state $\mathbb{S}_p$ based on the cells's neighborhood $N(p)$ and a defined local transition function $\delta: \mathbb{S}_p^t \rightarrow \mathbb{S}_p^{t+1}$. The cell's state evolves through discrete time steps $t$, where the state of a cell $p$ at time $t + 1$ depends on the configuration of states in its neighbors $N(p)$ at time $t$. Common neighborhoods are Von Neumann (4-neighbors) and Moore (8-neighbors).

CA applications on images range from classical image processing operations to more complex image analysis tasks. For example, in \cite{basic_ca_1}, they show the application of CA towards sharpening and smoothing operations, establishing that CA's methods can converge fast and are also stable in the presence of noise, hence proposing CA as a first-level elementary image enhancement technique. They also discuss the inherently parallel nature of CA and demonstrate results on 2D-Images comparable to widely used software back then, without requiring a priori information. Fast forwarding some years, \cite{basic_ca_2} further experimented with CA over additional operations --- noise filtering, thinning, and convex hulls --- applied on binary and grayscale images. Additionally, the main focus of their work was to draw a method to enable easy training CA, which means designing a set of rules to compose the transition function $\delta$. The authors discuss how even simple rules, combined with a matrix of cells with only local iterations, drive the system to map sophisticated global behaviors.

Then, building upon fundamental image processing techniques, other works have pushed CA applications to SOD. As a more complex task, SOD involves visual understanding and not only pixel-wise transformation (but local and object-level understanding). A seminal CA-SOD method is the work of \cite{Quin2015}, where they propose two CA as a post-processing operation capable of improving prior maps from all state-of-the-art methods back then. First, the authors argue that most SOD works take advantage of background priors, which means considering image boundaries as background. Despite this assumption holding for most images, the whole method will generate a wrong saliency when the assumption fails. Hence, to overcome this issue, they begin by segmenting the input image into N small super-pixels --- using the SLIC algorithm \citep{slic} --- described by their mean color features and coordinates. Through the K-means algorithm, boundary super-pixels are then clustered into $K$ clusters based on the CIE LAB color feature. Background seeds are computed from clustered super-pixels by combining geodesic and color information. Once background seeds are available, they evolve it using CA. Their approach considers each super-pixel as a cell, and each cell has as neighbors the adjacent super-pixels and cells with shared boundaries with the adjacent cells. An impact factor matrix and a coherence matrix then drive evolution, where the impact factor matrix measures the similarity in the CIE LAB color space between every cell, and the coherence matrix assesses the level of agreement between the state of the cell and its neighbors. As a last contribution, they propose a multi-level cellular automaton, which combines multiple state-of-the-art methods' saliency maps to generate a final saliency map. In this Multi-layer CA, each cell represents a pixel. Moreover, unlike the Single-layer CA, neighborhoods are defined across saliency maps rather than within a single map - specifically, pixels with the exact coordinates in different saliency maps are considered neighbors. They demonstrated results surpassing the current methods through a Bayesian framework, with their approach effectively integrating the complementary advantages of different SOD methods.

Additionally, \citeauthor{Quin2015} further contributed to the field by exploring a hierarchical CA towards SOD \citep{Qin2018}, again using a single- and a multi-level CA. This time, they extracted super-pixels for different image scales and deep features --- extracted using FCN-32s \citep{fcn_32} --- to guide evolution.

Semantic segmentation, another closely related image analysis domain where CA has been successfully applied, shares significant methodological overlap with SOD, where semantic segmentation is also concerned with classifying image pixels. However, it classifies every pixel/voxel between N classes (e.g., health tissue/tumor tissue), where SOD only concerns detecting the most salient object of the scene (.e.g., tumor tissue). Nevertheless, given the overlapping between the fields, we can drive inspiration from CA for semantic segmentation, mainly when applied to medical image analysis.

For example, in \cite{grow_cut}, the authors proposed a CA algorithm for semantic segmentation called GrowCut. Given an N-dimensional image as a lattice of cells, they describe each cell by a triple $(lb, \theta, \vec{C})$, where $lb$ is the label of the cell, $\theta$ the strength, and $\vec{C}$ the RGB vector. Then, the user initializes seed cells by adding markers on images with labels (e.g., background or foreground, or more classes); simultaneously, the cell's strength ($\theta$) is initialized. From these segmentation seeds, evolution happens, where the neighbors attack cells at each iteration. The attack force is controlled by the cell strength weighted by a monotonous decreasing function, which computes the distance between RGB vectors. If an adjacent cell has a higher strength value and is very similar to the central cell, it will conquer the central cell. Moreover, they also enable boundary smoothness by adding the concept of enemies (surrounding cells with different labels). Thresholding the number of enemies, they forbid cells to attack or conquer them automatically. Following the same trend, \cite{seeded_ca} explores seeded algorithms but embeds the Ford-Bellman algorithm as an evolution rule using GPU optimization. From $K$ labeled seeds, initialized by a user, they evolve the CA to compute multiple shortest-path trees. Once CA converges, $K$-cuts generate the final segmentation. Most interestingly, the authors integrate medical experts in the tests and show competitive performance (on low-cost graphics hardware suitable in 2010) for clinical routine. The possibility of implementing algorithms through CA is particularly interesting, as there is evidence that a CA with the proper rules can emulate a universal Turing machine and, therefore, implement anything computable \citep{cook2004universality}. Thus, it is possible to implement a more complex algorithm through carefully designed rules and parallelize it (given the characteristics of straightforward parallelization for CA).

The GrowCut algorithm was then adapted towards brain tumors in \cite{tumor_cut} through the Tumor-Cut algorithm. Considering the characteristic necrotic core of brain tumor MRI images, they propose a two-step method, where the whole tumor is first segmented. Then, the necrotic core is segmented through a CA-based approach. This time, user input must be a line drawn over the largest visible diameter of the tumor, which initializes cells with labels and strengths for the foreground (tumor). Additionally, it generates a volume of interest (VOI) over a sphere, which is 35\% larger than the drawn line. The VOI's border then initializes the background's labels and strengths. By evolving the foreground and background strengths, a tumor probability map is then calculated, which is used to evolve a level set surface to impose spatial smoothness, generating the final tumor segmentation. The necrotic core is segmented by computing two thresholds, one for necrotic voxels and the other for enhanced voxels. These thresholded voxels are initial seeds for a cellular automata algorithm that labels the remaining mid-intensity voxels. Years ago, the tumor cut algorithm was expanded to consider a second-order statistic (gray-level co-occurrence matrix) to evolve CA \citep{ca_glcm}.

Despite further modifications, all methods derived from GrowCut share the same evolving characteristic: a cell $q$ propagates its state to a cell $p$ at time $t + 1$ if both $p$ and $q$ cells are similar (close pixel color, intensity, or features) and if $q$ has more strength than $p$ given the monotonous decreasing function $g$, which means $g(p, q) \times \theta_{lb}^t (q) > \theta_{lb}^t (p)$.

\begin{figure*}[!htpb]
  \centering
  \includegraphics[width=\textwidth]{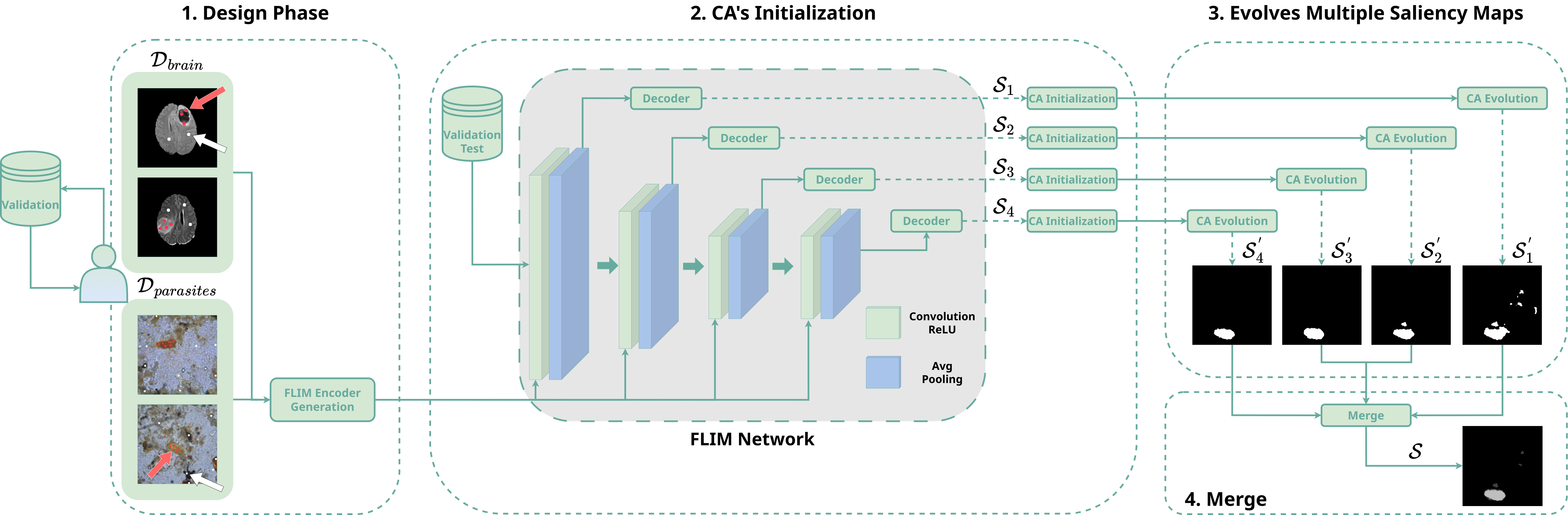}
  \caption{Multi-Level Cellular Automata for FLIM networks (example for a 4-layer FLIM Network): (1) Users iteratively select training images and define FLIM-Encoder architecture. (2) During inference, images enter the encoder, and features decode into intermediary saliency maps, initializing CA cells. (3) Multi-level saliency maps evolve from this initialization, and (4) merge into a final saliency.}
  \label{fig:method}
\end{figure*}

As a review of both SOD and segmentation methods reveals, CA enables applications on both natural and medical images. Nevertheless, most works initialize the CA by user interaction, which is acceptable for a few images but is burdensome, costly, and error-prone under scenarios where running for many images is mandatory. Given the works on FLIM previously discussed, the possibility of integrating FLIM networks with CA initialization arises. Interestingly, we can CA by employing previous user knowledge without requiring user interaction for every image. Hence, initializing CA employing a FLIM network is of particular interest.

\section{Multi-level Cellular Automata}
\label{sec:3}

Our method leverages the FLIM methodology to address the limitations of deep SOD literature. Figure \ref{fig:method} illustrates our proposed method implemented through a 4-level FLIM network, though the framework is adaptable to FLIM architectures with any number of layers ($L$). This section details each step of our method. First, we will approach the design phase, guiding the reader through selecting images and defining the architecture of a FLIM encoder. Then, we address how the decoded features are employed to initialize the CA's states. Once the multi-level CA are initialized, we will detail how they evolve and yield a probability map. Ultimately, we will detail an approach to merge multiple CAs' output.

\subsection{Design phase}

\begin{figure}[!ht]
    \centering
    \begin{subfigure}[b]{0.23\textwidth}
        \centering
        \includegraphics[width=\textwidth]{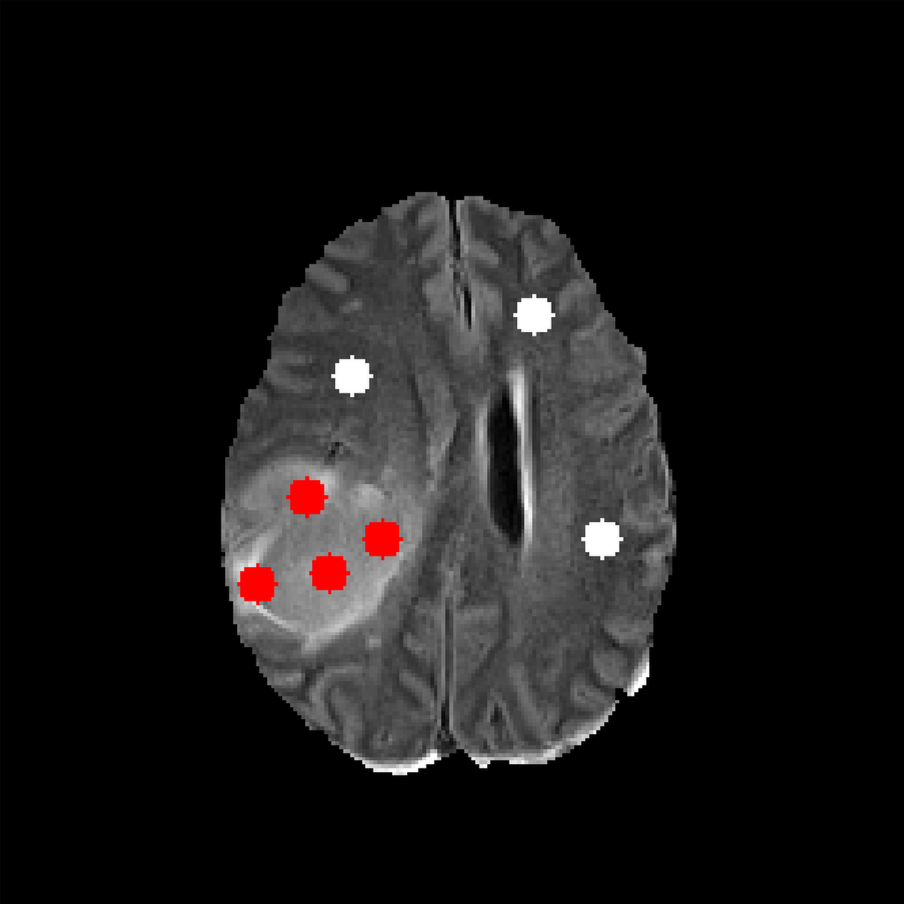}
        \caption{Markers on BraTS}
        \label{fig:image1}
    \end{subfigure}
    \begin{subfigure}[b]{0.23\textwidth}
        \centering
        \includegraphics[width=\textwidth]{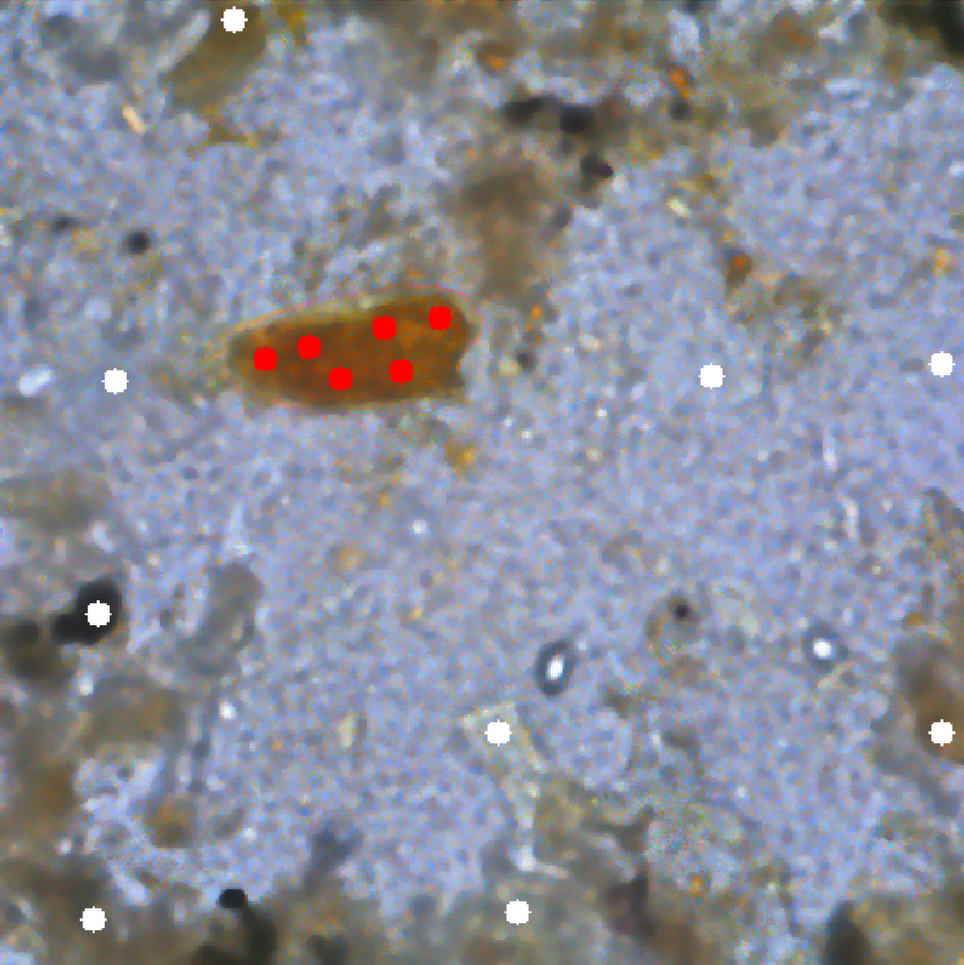}
        \caption{Markers on parasites}
        \label{fig:image2}
    \end{subfigure}
    \caption{Examples of user annotations: red markers indicate tumors or parasite eggs; white markers show healthy brain tissue, background, or debris.}
    \label{fig:sample_markers}
\end{figure}

The design phase focuses on training a FLIM Convolutional Encoder, the core component of the FLIM network. This first step is the only stage in our method that requires user interaction, which occurs in two specific, well-defined steps: First, as discussed in Section \ref{sec:flim_rw}, the user selects a set of representative images ($C_I$) following an iterative procedure. Second, given the selection of an initial image, the user places markers (dots) in descriptive and discriminative image regions. Figure \ref{fig:sample_markers} shows two examples of user makers inserted on selected images.

The designed FLIM encoder is then combined with the adaptive decoder (Equation \ref{eq:decoder}), which generates saliency maps for each network layer, such as $S_l = \{\mathcal{S}_l\}_{l=1}^L$, in which $L$ is the number of layers of the convolutional encoder. Then, the quality of saliency maps is assessed on a validation subset. The worst cases on validation may be selected to compose the training data, repeating the previous two steps. The iterative procedure is then stopped when acceptable performance is achieved, and this usually happens around 3 to 4 images for the evaluated datasets. Additionally, to supervise which image regions compose the convolutional filters, the user has complete control of the encoder's architecture through a configuration file.

\subsection{CA's initalization}
\label{sec:3_2}

Once we train the FLIM encoder, we can extract features at each encoder level by feeding it an input image. User interaction is only needed during the design phase. Now, the FLIM encoder is run in inference mode for remaining validation (excluding training images) and test images; validation is used to guide CA parametrization and test to assess our method. We then sequentially decode features at each level, generating an intermediate saliency map to initialize the state of the CA. Combining the FLIM encoder and adaptive decoders forms a \textit{FLIM network} \citep{gilson_sibgrapi_2024, crispim_sibgrapi_2024}, an architecture designed for extracting saliency maps. This $L$-layers network is developed end-to-end without requiring backpropagation and eliminates the need for pixel-wise annotated ground truths.

The CA's initialization corresponds to the second block in Figure 4, exemplified for a 4-layer FLIM network towards parasite egg detection; nevertheless, the method is suitable for encoders with any number of layers. We decode each encoder's features and generate an intermediary saliency map for each encoder level. Consequently, we generate a set of L intermediary saliency maps:  $S_l = \{\mathcal{S}_l\}_{l=1}^L$. Each intermediary saliency map initializes one CA, which represents a significant improvement over our previous work by leveraging the delineation-to-detection capability of FLIM-Networks (see Figures \ref{fig:grid_parasites_img2} to \ref{fig:grid_parasites_img5}). Using each FLIM network level, we can initialize multiple Cellular Automata (CAs)—one for each network layer—which expands our initialization space. This multi-level CA approach better captures both edge and internal regions of salient objects while effectively reducing false positives (see Figures \ref{fig:grid_parasites} and \ref{fig:grid_brain}).

To decode each layer feature map into an intermediary saliency map, we adopt a recently proposed adaptive decoder (Equation \ref{eq:w_adaptive}). Exploring decoders is outside the scope of this work, but we refer the reader to recent works that discuss decoders for FLIM networks ~\citep{joao2023flyweight, gilson_sibgrapi_2024, Joao2023BuildingFFVisApp}.

\begin{figure}[!ht]
    \centering
    \begin{subfigure}[b]{0.156\textwidth}
        \centering
        \includegraphics[width=\textwidth]{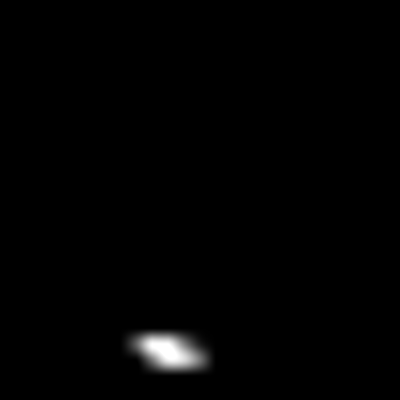}
        \caption{$S_4$}
        \label{fig:init_image1_parasites}
    \end{subfigure}
    \begin{subfigure}[b]{0.156\textwidth}
        \centering
        \includegraphics[width=\textwidth]{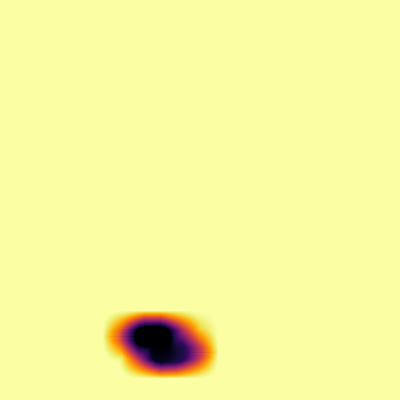}
        \caption{$\theta_0^0$}
        \label{fig:init_image2_parasites}
    \end{subfigure}
    \begin{subfigure}[b]{0.156\textwidth}
        \centering
        \includegraphics[width=\textwidth]{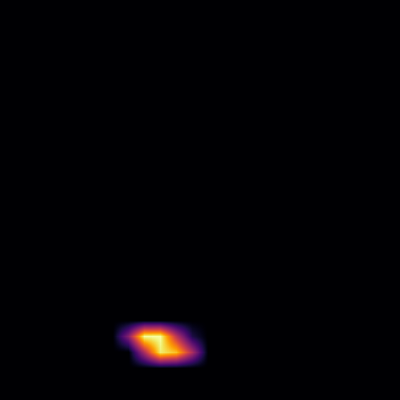}
        \caption{$\theta_1^0$}
        \label{fig:init_image3_parasites}
    \end{subfigure}
    \caption{Example of initialization procedure for parasite eggs detection. FLIM-Network saliency from layer 4 (a) initializes background (b) and foreground strength (c).}
    \label{fig:initialization_parasites}
\end{figure}

From the saliency map of each layer $S_l = \{\mathcal{S}_l\}_{l=1}^L$, we initialize the CA cell's states, where $\theta_0^0$ and $\theta_1^0$ represents background and foreground strengths, respectively, at $t = 0$. For $\theta_1^0$, we simply normalize $S_l$ within $[0, 1]$ and set cells' foreground strengths $\theta_1^0(p)$. However, the FLIM encoder may have pooling operations with strides larger than 1 for both datasets. If necessary, we upsample the saliency map $S_l$ (bilinear upsampling) to the same size as the input image, as image color/intensity are employed to guide CA's evolution, and the loss of edge information through downsampling operations may degrade CA's performance.

Initializing the background strengths, $\theta_0^0$, is more challenging, and hence our method considers two approaches given the dataset at hand. The first approach assumes that FLIM networks are good at detection and that the CA's can improve salience delineation while eliminating false positives. Hence, we initialize $\theta_0^0$ as the complement of saliency maps ($S_l$) posterior to a dilation operation of radius 10. Accordingly, we are abble to evaluate CA's capability of correctly evolving to fit tumor and parasite eggs. Figure \ref{fig:initialization_parasites} shows an example where no prior assumption can be made about where the salient object (parasite eggs) may appear. For example, background priors \citep{Qin2018} are unsuitable, as eggs can be spotted at image borders.

The second approach concerns applications where a reasonable assumption about the background can be made. In such cases, we initialize the background strength as $\theta_0^0 \leftarrow 1$ based on a mask. Figure \ref{fig:initialization_parasites} exemplifies such an example, where the salient object is brain tumors. As brain tumors only appear inside the brain, we can employ a \textit{brain prior}, were given a brain mask, background strengths are easily defined.

\begin{figure}[!ht]
    \centering
    \begin{subfigure}[b]{0.156\textwidth}
        \centering
        \includegraphics[width=\textwidth]{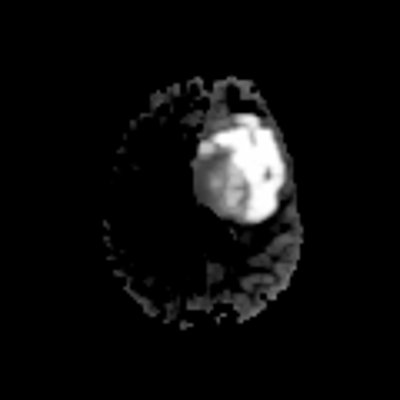}
        \caption{$S_1$}
        \label{fig:init_image1_brain}
    \end{subfigure}
    \begin{subfigure}[b]{0.156\textwidth}
        \centering
        \includegraphics[width=\textwidth]{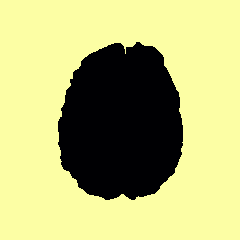}
        \caption{$\theta_0^0$}
        \label{fig:init_image2_brain}
    \end{subfigure}
    \begin{subfigure}[b]{0.156\textwidth}
        \centering
        \includegraphics[width=\textwidth]{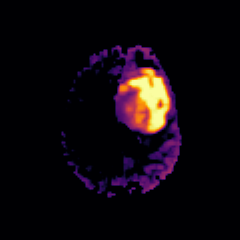}
        \caption{$\theta_1^0$}
        \label{fig:init_image3_brain}
    \end{subfigure}
    \caption{Example of the initialization procedure for brain tumor detection. FLIM-Network saliency from layer 1 (a) initializes background (b) and foreground strength (c).}
    \label{fig:initialization_brain}
\end{figure}

Lastly, we need to initialize the label map $LM$ at $t = 0$, where we set $LM(p) \leftarrow 1$, if $\theta_{1}^0(p) > 0$, and $0$ otherwise. The described initialization procedure is the same as that applied to each layer, and once it is done, we evolve each CA (one for each encoder level).

\subsection{CA's evolution}

The CA's evolution is detailed in Algorithm \ref{alg:ca}, where the same algorithm details the evolution of each CA. Hence, what we describe here is the same as that applied to each CA; the only difference is how they were initialized. Once the cell's state is set, we evolve each CA state (\ie foreground and background) until convergence is reached --- when there is little to no update on a given evolution step.

\begin{algorithm}[!ht]
    \caption{Cellular Automata initialized by FLIM}\label{alg:ca} % ca_intensity
    \begin{algorithmic}[1]   
        \Statex \textbf{Input:}
        \Statex \hspace{\algorithmicindent} $\mathbf{I}$ - input image, where $\hat{\mathbf{I}} = (D, \mathbf{I})$;
        \Statex \hspace{\algorithmicindent} $\theta$ - Cells strength, initialized by FLIM, where $\hat{\theta} = (D, \theta)$;
        \Statex \hspace{\algorithmicindent} $ol$ - object label being evolved (0 | 1, background | foreground);
        \Statex \hspace{\algorithmicindent} $LM$ - Cells' current label, where $\hat{LM} = (D, LM)$.
        \Statex \textbf{Output:} Evolved Cells' strength.
        \Procedure{CA}{$\mathbf{I}$, $\theta_{ol}^0, ol, LM$}
            \State $t \gets 0$
            \State $dist \gets +\infty$
            \While{ $ dist > 10^{-8}$ }
                \State $\theta_{ol}^{t + 1} \gets \theta_{ol}^{t}$ \Comment{Copy previous $\theta$}
                \For{\texttt{$\forall p \in D$}}
                    \State $q_{max} \gets \theta_{l}^t(p)$
                    \For{\texttt{$\forall q \in N(p)$}} \Comment{Moore}
                        \State $q_{aux} = g(p, q) \times \theta_{ol}^t(q)$
                        \If{$q_{aux} > q_{max}$}
                            \State $\theta_{ol}^{t + 1}(p) \gets q_{aux}$
                            \State $LM (p) \gets LM (q)$
                            \State $q_{max} \gets q_{aux}$
                        \EndIf
                    \EndFor
                \EndFor
                \State $dist \gets ||\theta_{ol}^t - \theta_{ol}^{t+1}||_2 / |D_{I}|$
                \State $\theta_{ol}^t \gets \theta_{ol}^{t+1}$ \Comment{Update step}
                \State $t \gets t + 1$
            \EndWhile
        \State \Return $\theta_l^t$
        \EndProcedure
    \end{algorithmic}
\end{algorithm}

The evolution happens based on a Moore Neighborhood (8-neighbors), where we took inspiration from the tumor cut algorithm proposed by \cite{tumor_cut}. We evolve for each object label, $ol \in \{0, 1\}$, background and foreground, respectively. The execution occurs first for $ol=1$ (foreground) and later for $ol=0$ (background). 

Given an input image $I$, initial strength $\theta_{ol}^0$, and label map $LM$, the algorithm begins by initializing a control variable $dist$ (Line 3) used for convergence detection in the main loop (Line 3). In each iteration, the future cell's strength $\theta_{ol}^{t+1}$ is copied from the current strength (Line 5) to keep track of unconquered cells.

The core evolution process (Lines 6-16) evolves each cell $p$ in the domain. For each cell, the algorithm sets the $q_{max}$ as the current strength of $p$ (Line 7) and then evaluates all neighboring pixels $q$ in the Moore neighborhood (Line 8). For each neighbor, it calculates $q_{aux}$ as the product of a similarity function $g(p,q)$ and the neighbor's current strength $\theta_l^t(q)$ (Line 9). If this calculated value exceeds the current maximum strength $q_{max}$ (Line 9), the cell $p$ is conquered (Lines 11-13).

After processing all pixels, the algorithm computes the L2-norm between current and future strengths normalized by domain size to measure convergence (Line 17). The current strength $\theta_l^t$ is updated with the future strength (Line 18), and the iteration counter $t$ is incremented (Line 19). This process continues until the convergence is reached ($dist $ distance falls below $10^{-8}$, indicating convergence.

The similarity function $g(p,q)$ is computed according to Equation \ref{eq:similarity_function}, which incorporates image features (intensity/color) to guide the evolution process. The $\beta$ parameter smooths the weight reduction during evolution if evolving inside the object (see Section \ref{sec:4_2}), which means that a lower value will cause foreground regions to leak. In contrast, higher values will penalize the saliency map on edge regions, as cells will lose strength fast and may not fit correctly to the object border.

\begin{equation}
g(p, q) = \begin{cases}
    e^{\beta ||\vec{I}(p) - \vec{I}(q)||_2}, & \mbox{\scalefont{0.9}if evolving object} \\
    e^{||\vec{I}(p) - \vec{I}(q)||_2},      & \mbox{\scalefont{0.9} otherwise},
\end{cases} \label{eq:similarity_function}
\end{equation}

\begin{figure}[!ht]
    \centering 
    \includegraphics[width=0.48\textwidth]{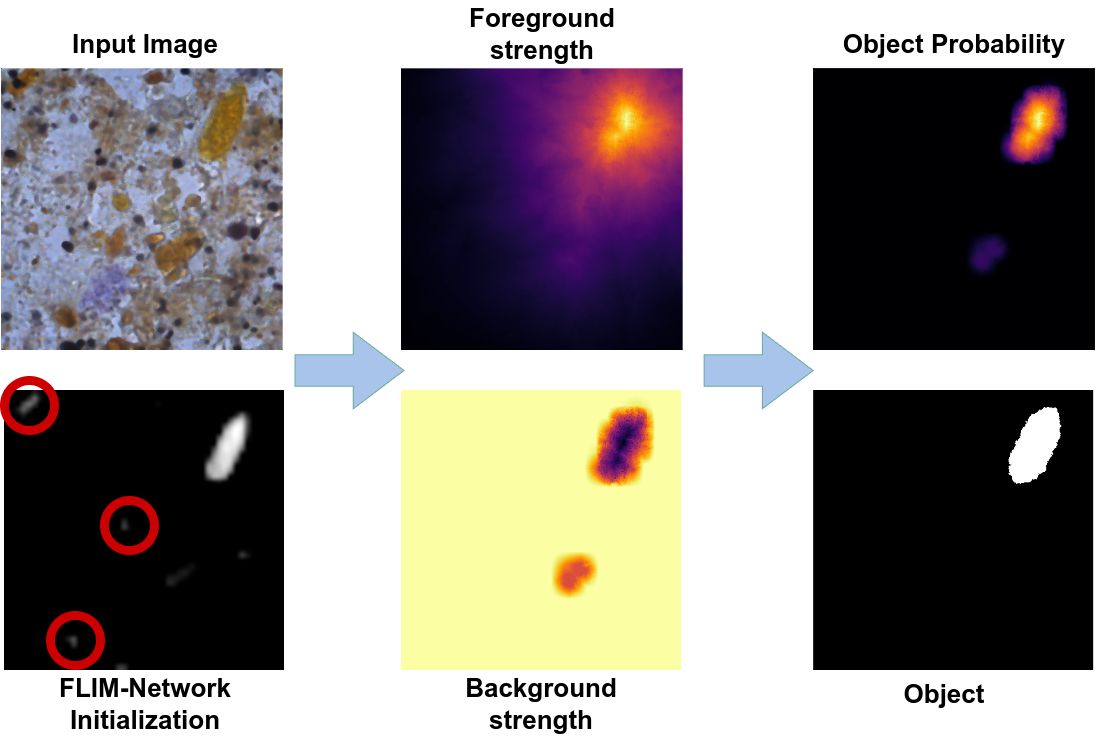}
    \caption{The first column shows the input image and FLIM initialization to CA. The middle column shows the evolved foreground and background strengths. The last column shows the object probability map and binary saliency. Our method evolves correctly even with false positives (red circles).}
\label{fig:ca_ev_parasites}
\end{figure}

Once convergence is reached, combining foreground ($\theta_{1}^{t}$) and background ($\theta_{0}^{t}$) strengths generates a object probability map $O_l$ (Equation \ref{eq:O}), and we define a binary saliency map $S'_l$ by thresholding $O_l$. Figure \ref{fig:ca_ev_parasites} exemplifies the whole evolution pipeline, where from FLIM-Network initialization, we evolve the foreground and background strengths (middle column), which enables us to compute the object probability map ($O_l$) and thresholding the parasite egg.

\begin{equation}\label{eq:O}
O_l (p) = \frac{ln(\theta_{0}^t(p))}{ln(\theta_{0}^t(p)) + ln(\theta_{1}^t(p))} 
\end{equation}

As exemplified in Figure \ref{fig:method}, the CA evolution is performed for each FLIM-Network level, producing $L$ saliency maps, which explores multi-level hierarchical features. Through multiple initializations, we now have an ensemble of complementary saliency maps that can be combined to produce a more robust final segmentation result.

\subsection{Merge multiple saliency maps}

\begin{figure}[!ht]
    \centering 
    \includegraphics[width=0.48\textwidth]{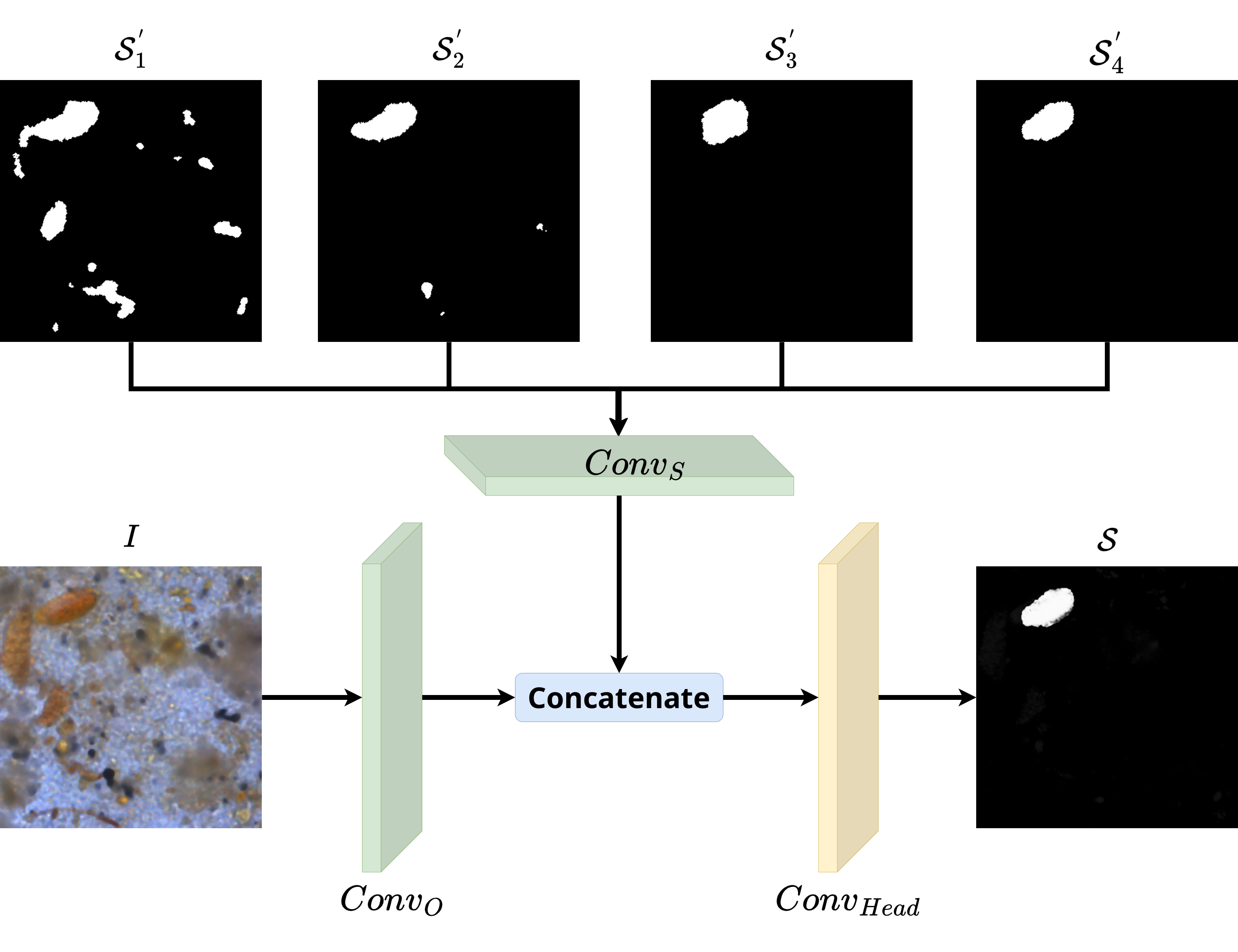}
    \caption{Merging multiple saliency maps from CAs supervised by input image.}
\label{fig:merge_network}
\end{figure}

Given the improved saliency maps from CA Evolution, we want to merge them into a single final salience map. As each evolved saliency corresponds to a hierarchical layer in our FLIM network, we want to leverage the best of each saliency without increasing the method's complexity. The inputs of the merging network are all evolved saliency maps $S'_l = \{\mathcal{S}'_l\}_{l=1}^L$ and the input image, which guides the merge procedure. During the initialization phase, if necessary, we upsample the saliencies to the size of the input image; then, the evolved saliencies are the same size. We designed a straightforward and efficient fully convolutional neural network to merge all evolved saliency maps into a final saliency map (guided by the input image). Figures \ref{fig:merge_network} show an example of such a merge network for a 4-layer FLIM network.

The simplicity of this merging network is well suited to our problem, where we face two main challenges: learning from few images (3-4) and learning from imperfect saliency maps. We want to combine multiple-level saliency maps, merging correct regions and eliminating incorrect ones. We train this simple model for 2000 epochs, with a learning rate of (1e-2), Adam optimizer, and l1 regularization (with $\lambda = 1e-3)$; we also employed the cosine annealing scheduler to take advantage of lower learning rates and learning resets. The key aspect of our training procedure was aggressive augmentations, where we employed random crop, horizontal and vertical flip, rotation, sharpness, affine, and perspective transformations. Our implementation guarantees that the same augmentations are applied to the multiple-saliencies and input image.

The network is composed of 3 convolutional filters. First, a single $3 \times 3 \times c$ kernel ($Conv_O$), in which $c$ is the number of channels of the input image (\eg 3 for RGB images and 1 for gray-scale ones), extracts features from the input image ($I$). In parallel, a single $3 \times 3 \times L$ kernel ($Conv_S$) combines the $L$ saliency maps. Both convolutions generate a single-channel feature map, to which we apply a sigmoid activation function. The two feature maps are concatenated and fed to $Conv_{head}$, which generates the final saliency map with a point-wise convolution of $1 \times 1 \times 2$ followed by a sigmoid.

At this stage, our approach resembles an ensemble of CAs initialized through multi-level features, namely a multi-level CA. We can replicate the proposed method for multi-channel images, focusing on scenarios where collecting large datasets and pixel-wisely annotating images is problematic.

\section{Experimental setup}
\label{sec:4}

This section details the setup to validate the proposed method on two challenging datasets. We will guide the reader through building a FLIM network and its integration into the multi-level CA approach, discussing how to initialize, evolve, and merge the saliency maps. Additionally, this section also presents the SOD models used for comparison with our method, as well as the evaluation metrics employed for assessment.

\subsection{Designing a FLIM Network}

We choose two medical datasets to validate the proposed method: the BraTS 2021 \citep{brats2021} for brain tumor segmentation and a private dataset of parasite eggs. For BraTS, we extract 2D axial slices of brain tumors (T2-FLAIR), three slices for every image (given the tumor volume, we select axial indexes representing the median, $1^{st}$ and $3^{tr}$ quartile). For parasites, images were collected through a method (DAPI) developed in our lab focused on the automated diagnosis of human and pet parasites ~\citep{SuzukiTBE2013}.

The problem of parasites is intrinsically related to the deep SOD problems, mainly inference time and annotation cost. Given an input slide, DAPI performs the screening and processing of around 2000 images with four million pixels each, segmenting and identifying objects of interest, where the whole pipeline must take less than 5 minutes. Recent works show that deep neural networks improve the diagnosis of gastrointestinal parasites in humans, cats, and dogs ~\citep{OsakuCBM2020, JoaoCBM2023}. However, given the need for less than 5-minute processing times, applying those methods without considerably increasing costs is impossible. Moreover, annotating the volume of ground-truth data required for training is both timely and financially expensive. Hence, we need lighter networks, and at the same time, we want to take advantage of weak annotations.

With the goal of broader validation of our method, we chose to explore the BraTS 2D dataset, as it enables us to assess our image in a different domain (MRI images), with grayscale images, and with a more heterogeneous dataset (multi-institutional dataset).

Once datasets are prepared, we split them into three splits, each composed of validation and test subsets, following a 50/50 ratio. For parasites, after the splitting, we also filter only images that contain parasites. Hence, we end up with the following configurations of splits: 336/296, 308/324, and 326/306 (validation/test images). For the brain tumor dataset, the three splits have 1882/1882 validation and test images, as we guarantee during dataset construction that all images hold tumor regions.

The next step is to generate the architecture of the FLIM Encoder. As we previously detailed in Section \ref{sec:flim_rw}, there are two steps where the FLIM methodology performs filter reduction and, hence, harnesses user control. First, when the number of filters extracted from one image (from all markers) is larger than the upper limit of filters set for each image, and second, if the total number of filters learned is larger than the number of filters specified in the convolutional layer. To avoid those reductions, we took the following decisions:

\begin{enumerate}
    \item We only use disks as marks (\ie user clicks, see Figure \ref{fig:sample_markers}), so we control the exact position that generates filters;
    \item Each marker generates four filters;
    \item The specified number of filters for each convolutional layer is 200, and given the number of selected images and drawn markers, we never surpass this number.
\end{enumerate}

We proceed to image selection after specifying these requirements in the FLIM network. First, one image is randomly chosen from the validation subset, and we draw markers on descriptive and discriminative regions (Figure \ref{fig:sample_markers} exemplifies those markers). One can see this procedure as weak supervision once there is no need for a densely annotated ground truth, and filters are learned directly from marked regions. After a FLIM encoder is created, we apply the adaptive decoder described in Equation \ref{eq:decoder}, which is empirically configured with $A_1 = 0.1$ and $A_2 = 0.2$ for the area thresholds. Then, we compute metrics (F-Score, Dice, MAE) for all validation data and select a new image for annotation based on the worst predictions, increasing the training subset. We repeat this procedure until the desired performance is achieved. It is a manual procedure, but there is parallel work on automating image selection for training \citep{abrantes_sibgrapi_2024, abrantes_embc}.

\begin{table}[htbp]
\centering
\caption{FLIM Encoder for Parasites dataset (s for stride)}
\begin{tabular}{|c|c|c|c|}
\hline
\textbf{Layer} & \textbf{Conv} & \textbf{Activation} & \textbf{Pooling} \\
\hline
1 &  $(3 \times 3)$ & ReLU & Avg $(3 \times 3 | s2)$ \\
\hline
2 &  $(3 \times 3)$ & ReLU & Avg $(3 \times 3 | s2)$ \\
\hline
3 &  $(3 \times 3)$ & ReLU & Avg $(3 \times 3 | s1)$ \\
\hline
4 &  $(3 \times 3)$ & ReLU & Avg $(3 \times 3 | s1)$ \\
\hline
\end{tabular}
\label{tab:architecture_parasite}
\end{table}

A suitable architecture for CA initialization is typically achieved within 3 to 4 training images. Table \ref{tab:architecture_parasite} presents the encoder architecture for parasites, while Table \ref{tab:architecture_BraTS} displays the 2D BraTS encoder architecture.

\begin{table}[htbp]
\centering
\caption{FLIM Encoder for BraTS 2D dataset (s for stride)}
\begin{tabular}{|c|c|c|c|}
\hline
\textbf{Layer} & \textbf{Conv} & \textbf{Activation} & \textbf{Pooling} \\
\hline
1 &  $(3 \times 3)$ & ReLU & Max $(3 \times 3 | s2)$ \\
\hline
2 &  $(3 \times 3)$ & ReLU & Max $(3 \times 3 | s1)$ \\
\hline
3 &  $(3 \times 3)$ & ReLU & Max $(3 \times 3 | s1)$ \\
\hline
\end{tabular}
\label{tab:architecture_BraTS}
\end{table}

As the number of filters depends on the number of markers on training images, they differ slightly across splits. On parasites, we have 121, 147, and 141 filters for the three splits, respectively. The number of filters across splits for brain tumors is 60, 48, and 68. All layers within each split maintain the same number of filters for both datasets.

We develop the FLIM-Encoder architectures empirically, based upon an experimental analysis. A deeper study of designing encoders is outside our scope since our goal was to validate multi-level CA initialization through FLIM. Once this step is finished, we have a convolutional encoder (FLIM-based) ready for inference and initializing CA's cell states. By combining these encoder with our adaptive decoder, the same across all feature levels, we have a FLIM network.

The FLIM network consists of different numbers of layers for each dataset. Our method is evaluated using a 4-layer FLIM network for parasites and a 3-layer FLIM network for brain tumors. For the parasites, the saliency maps are represented as \([S_1, S_2, S_3, S_4]\), as illustrated in the method pipeline. For brain tumors, we generate three saliency maps \([S_1, S_2, S_3]\) from the 3-layer convolutional encoder.

\subsection{Multi-level CA parametrization and execution}

Given the two defined datasets and the extracted saliency maps, both datasets' foreground strengths are initialized identically. However, as depicted in Section \ref{sec:3_2}, the brain dataset allows us to initialize the background strength based on a prior (\textit{brain prior}), which assumes that background strengths are the maximum value outside the brain. Then, for BraTS, we used the brain mask, and the background strengths are easily defined as $\theta_0^0 \leftarrow 0$ (inside the brain) and $\theta_0^0 \leftarrow 1$ (outside the brain). For parasites, posterior to a dilation operation of radius 10. For initialization examples, see Figures \ref{fig:initialization_parasites} and \ref{fig:initialization_brain}.

After all CA are initialized, we evolve them according to the Algorithm \ref{alg:ca}. To experiment with CA algorithm parametrization, we explored the validation data, and once we had selected its parameters (i.e., different thresholding on brain images), we used our test subsets to assess our results. Since we are dealing with gray-scale and RGB images, the first configuration specific for each dataset is the $\beta$ parameter in Equation \ref{eq:similarity_function}. For the brain, we smooth evolution when evolving in the foreground ($LM(q) = 1$), and if $I_1(p) > I_1(q)$ (gray-scale images), given the fact that the tumor core is darker than the surrounding tumor (edematous tissue). For parasites, we also verify if evolving on the foreground and the Euclidean distance in LAB color space ($\|\vec{I}(p) - \vec{I}(q)\|_2 < 0.2$). For both datasets, when smoothing $\beta=0.6$. 

Once evolution converged, we now have multiple object probability maps, $O_l = \{\mathcal{O}_l\}_{l=1}^L$, which must be thresholded to generate the improved saliency maps $S'_l = \{\mathcal{S}'_l\}_{l=1}^L$. Thresholding the object probability map required special attention. For parasites, $S'_l$ is defined by pixels with $O_l(p) > Otsu(O_l)$ (Figure \ref{fig:ca_ev_parasites} exemplifies the evolution end-to-end). For brain tumors, a different approach was required. Rather than using Otsu thresholding on the probability map, a histogram-based statistical analysis performs better. We first analyze pixel intensity distributions within the brain mask, focusing on the upper-intensity spectrum (top 10\%), which enhances the detection of the characteristic peak of the whole tumor on T2-FLAIR images. After identifying this peak intensity, we compute the mean ($\mu$) and standard deviation ($\sigma$) of intensity values within a 20\% window around the peak. The final threshold is then defined as $\tau = \mu - k\sigma$, where $k$ is a constant parameter (empirically set to $k=2.5$). Any pixel with intensity greater than this threshold ($I(p) \geq \tau$) is classified as tumor tissue in the binary saliency map $S'_l$. This statistical approach better accommodates the distinctive high-intensity characteristics of tumor regions T2-FLAIR images while adaptively adjusting to image-specific intensity distributions.

Our final step in multi-level CA now concerns merging the improved saliency maps $ S'_l = \{\mathcal{S}'_l\}_{l=1}^L$. As shown in Figure \ref{fig:method}, the CA evolution is performed for each FLIM-Network level, producing four saliency maps for parasite segmentation and three saliency maps for brain tumor detection, exploring multi-level hierarchical features, which now composes an ensemble of CA, or the multi-level CA, through the merging network. Through multiple initializations, we now have an ensemble of complementary saliency maps that can be combined to produce a more robust final segmentation result. The only difference between the merging network for the two chosen datasets happens in the input filters, where the input convolution has $c=3$ (RGB) and $L=4$ for parasites and $c=1$ (gray-scale) and $L=3$ for brain tumors.

\subsection{Evaluation Metrics \& Benchmark}
\label{sec:4_2}

To evaluate the efficacy of multi-level CA for FLIM-Networks, we employed the following metrics: F-Score, $\mu$WF, Dice, e-measure, s-measure, and MAE. All metrics, except for MAE, range from 0\% to 100\%, with higher values indicating better saliency maps.

The F-Score and $\mu$WF metrics assess the quality of the binarized saliency map (threshold set as 0.5) by combining precision and recall \citep{fscore, meamwf}. Following the literature, we empirically set $\beta^2 = 0.3$ to give more emphasis on precision. Similarly, Dice measures the overlapping ratio between the binary saliency map and the binary ground truth, making it particularly effective for evaluating improvements in edge regions \citep{basnet}. The s-measure \citep{smeasure} evaluates the structural similarity between the normalized saliency map and the binary ground truth (with $\alpha=0.5$). After subtracting their global means, the e-measure \citep{emeasure} quantifies the correlation between the binary saliency map and ground truth. The e-measure metric simultaneously accounts for global statistics and local pixel-level matching. MAE \citep{mae} computes the pixel-wise absolute difference between the normalized saliency map and the binary ground-truth mask, directly measuring prediction deviation. Before metrics computing, we filter components by area: $[1000, 9000]$ for parasites and $[100, 20000]$ for brain.

Leveraging the metrics above, we then compare our model to three recent lightweight models: SAMNet \citep{liu2021samnet}, MSCNet \citep{mscnet}, and MEANet \citep{liang2023meanet}; and to two recent heavier models: BasNet \citep{basnet} and U$\mathbf{^2}$-Net \citep{u2net}. Our goal is to benchmark our model against deep SOD, comparing the performance of those models under constrained scenarios where annotated data are scarce. Therefore, we fine-tune those models on the same training images employed in the FLIM design phase (3-4 images). Remarkably, FLIM networks are significantly smaller than lightweight networks.

To fine-tune the models, we utilized the weights released by the authors. For BasNet and U$\mathbf{^2}$, trained on the augmented DUTs-TR dataset (21,106 images), we fine-tuned with a lower learning rate (1e-4) and additional augmentations (vertical/horizontal flips, rotation, color jittering, Gaussian noise). For lightweight models, we used the authors' pretrained models and code, where even the backbone had been previously fine-tuned on ImageNet.

\section{Results \& Discussion}
\label{sec:5}

In this study, we present an investigation of a novel initialization method for CA while also exploring an ensemble of multi-level CAs for two datasets: parasite eggs and brain tumors. We want a saliency map on parasites that correctly locates and delineates a parasite egg. As we process T2-FLAIR images for brain tumors, we aim to detect and delineate the whole tumor (a hyperintense signal on T2-FLAIR sequences).	

\begin{figure}[!ht]
    \centering 
    \includegraphics[width=0.48\textwidth]{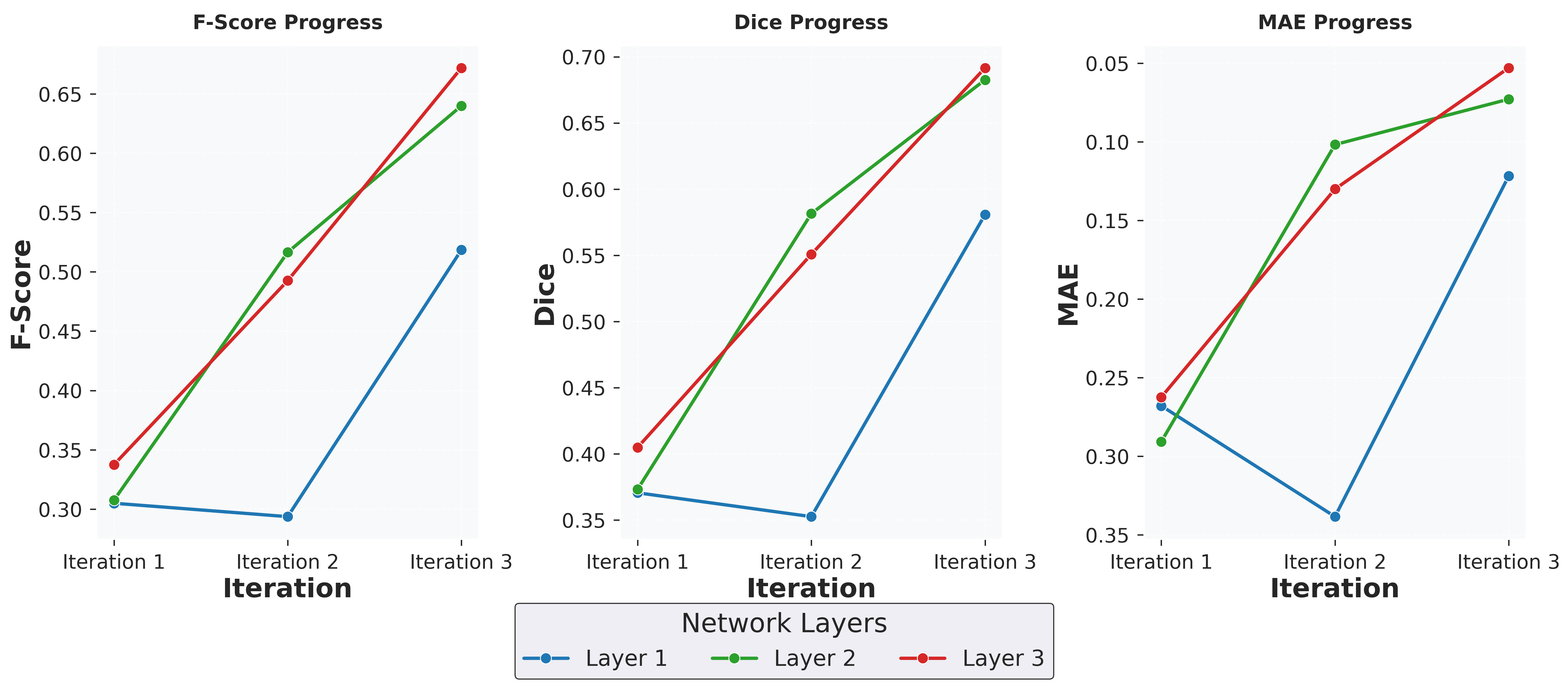}
    \caption{Metrics improvement on validation data across FLIM design. We start with one training image and at each iteration we add one more iumage.}
\label{fig:graph_img_selection}
\end{figure}

The first step in enabling the proposed method involves selecting a random image to train an initial FLIM encoder. By decoding extracted features, we can compute metrics for all validation data, where images with bad performance are candidates for increasing the training data. Figure \ref{fig:graph_img_selection} exemplifies this procedure for building the FLIM encoder for brain tumor split 1. We see significant improvements over the analyzed metrics across each iteration. For instance, at iteration 1, with one image, F-Score values are around 0.3 to 035 for layers 1 to 3. However, as we add one more image training image, there is a slight decrease in F-Score for layer 1, while we see an increase of around 0.15 for layers 2 and 3. An improvement for all layers is observable once we add a third training image. Since layers 2 and 3 show metric values around 0.65 at this iteration, we assume the FLIM encoder is suitable for CA initialization, and the training procedure is done. This iterative procedure has two key benefits:

\begin{enumerate}
    \item It allows choosing the right amount of training data;
    \item It also allows the identification of the suitable network architecture for the given problem. For instance, if the performance of layer 3 is worse than that of layer 2, we can modify our FLIM encoder to have only two convolutional layers.
\end{enumerate}

Once the design phase is done, we will have a FLIM network, and we can run inferences. We first extract each validation and test image's features, decode them into an intermediary saliency map (one for each encoder level), and then initialize our multiple CAs, where evolutions happen until convergence. Tables \ref{tab:parasites_all} and \ref{tab:brain_all} summarize our results on test sets (average and standard deviation across splits) for parasites and BraTS 2d, respectively.

Table \ref{tab:parasites_all} demonstrates that multi-level CA significantly enhances saliency across all layers of parasites. Comparing intermediary saliency to CA output, we observed improvements in F-Score (by 2.8-13.7\%), $\mu$WF (by 26.7-43.4\%), Dice coefficient (by 26.2-41.3\%), e-measure (by 22.4-34.4\%), and s-measure (by 28.1-37.4\%). For MAE, our method reduced error values by 0.091-0.31, significantly improving pixel-wise accuracy. These results confirm that our approach enhances precision, recall, delineation quality, structural similarity, and correlation between predictions and ground-truth masks while reducing overall error. Our method exhibits greater consistency, as demonstrated by lower standard deviation values across all metrics.

The improvements across all layers yield suitable saliency maps for merging, which can be characterized as a Multi-Level CA or a CA's ensemble. On the merged column on Table \ref{tab:parasites_all}, we visualize that by a simple network composed of 3 convolutional filters, we can generate a saliency map that combines multiple saliencies into a better one. Except for MAE, the merged saliency surpasses metrics on all layers. Nevertheless, even for MAE, the average metric is closer to the better layer (layers 2 and 3).

\begin{table*}[!ht]
\caption{Metrics for parasites on test.} 
\centering
\begin{tabular*}{\textwidth}{@{}c*{6}{@{\extracolsep{\fill}}c}@{}}
\hline \hline
 \textbf{Metric} & \textbf{Model} & \textbf{Layer\_1}   & \textbf{Layer\_2}     &  \textbf{Layer\_3}  &  \textbf{Layer\_4} &  \textbf{Merged}\\
\hline \hline
\multirow{2}{*}{\begin{tabular}[c]{@{}c@{}}\textbf{F-Score}\\\textbf{(}$\mathbf{\uparrow}$\textbf{)}\end{tabular}} 
& FLIM & 0.467 $\pm$ 0.053 & 0.690 $\pm$ 0.022 & 0.726 $\pm$ 0.020 & 0.602 $\pm$ 0.027 & 0.821 $\pm$ 0.010 \\
& FLIM\_MCA & 0.604 $\pm$ 0.019 & 0.718 $\pm$ 0.020 & 0.780 $\pm$ 0.011 & 0.779 $\pm$ 0.011 & 0.834 $\pm$ 0.017 \\
\hline
\multirow{2}{*}{\begin{tabular}[c]{@{}c@{}}$\mathbf{\mu}$\textbf{WF}\\\textbf{(}$\mathbf{\uparrow}$\textbf{)}\end{tabular}}
& FLIM & 0.310 $\pm$ 0.042 & 0.472 $\pm$ 0.019 & 0.480 $\pm$ 0.019 & 0.350 $\pm$ 0.018 & 0.660 $\pm$ 0.011 \\
& FLIM\_MCA & 0.632 $\pm$ 0.021 & 0.739 $\pm$ 0.017 & 0.795 $\pm$ 0.009 & 0.784 $\pm$ 0.011 & 0.816 $\pm$ 0.017 \\
\hline
\multirow{2}{*}{\begin{tabular}[c]{@{}c@{}}\textbf{Dice}\\\textbf{(}$\mathbf{\uparrow}$\textbf{)}\end{tabular}} 
& FLIM & 0.322 $\pm$ 0.043 & 0.494 $\pm$ 0.020 & 0.505 $\pm$ 0.020 & 0.378 $\pm$ 0.019 & 0.680 $\pm$ 0.010 \\
& FLIM\_MCA & 0.657 $\pm$ 0.021 & 0.756 $\pm$ 0.015 & 0.805 $\pm$ 0.007 & 0.791 $\pm$ 0.009 & 0.823 $\pm$ 0.015 \\
\hline
\multirow{2}{*}{\begin{tabular}[c]{@{}c@{}}\textbf{e-measure}\\\textbf{(}$\mathbf{\uparrow}$\textbf{)}\end{tabular}}
& FLIM & 0.518 $\pm$ 0.058 & 0.688 $\pm$ 0.015 & 0.701 $\pm$ 0.011 & 0.598 $\pm$ 0.013 & 0.848 $\pm$ 0.014 \\
& FLIM\_MCA & 0.862 $\pm$ 0.005 & 0.912 $\pm$ 0.007 & 0.935 $\pm$ 0.007 & 0.936 $\pm$ 0.008 & 0.940 $\pm$ 0.004 \\
\hline
\multirow{2}{*}{\begin{tabular}[c]{@{}c@{}}\textbf{s-measure}\\\textbf{(}$\mathbf{\uparrow}$\textbf{)}\end{tabular}}
& FLIM & 0.404 $\pm$ 0.043 & 0.558 $\pm$ 0.014 & 0.584 $\pm$ 0.011 & 0.517 $\pm$ 0.012 & 0.756 $\pm$ 0.012 \\
& FLIM\_MCA & 0.778 $\pm$ 0.010 & 0.840 $\pm$ 0.009 & 0.867 $\pm$ 0.006 & 0.850 $\pm$ 0.006 & 0.879 $\pm$ 0.008 \\
\hline
\multirow{2}{*}{\begin{tabular}[c]{@{}c@{}}\textbf{MAE}\\\textbf{(}$\mathbf{\downarrow}$\textbf{)}\end{tabular}}
& FLIM & 0.337 $\pm$ 0.061 & 0.146 $\pm$ 0.013 & 0.110 $\pm$ 0.006 & 0.168 $\pm$ 0.014 & 0.057 $\pm$ 0.007 \\
& FLIM\_MCA & 0.030 $\pm$ 0.001 & 0.019 $\pm$ 0.001 & 0.019 $\pm$ 0.002 & 0.032 $\pm$ 0.003 & 0.031 $\pm$ 0.008 \\
\hline \hline
\end{tabular*}\label{tab:parasites_all}
\end{table*}

\begin{table*}[!htb]
\caption{Benchmark on parasites test set (FLIM Metrics stands for merged model).}
\centering
\begin{tabular*}{\textwidth}{@{}c*{6}{@{\extracolsep{\fill}}c}@{}}
\hline \hline
\textbf{Model} & \textbf{F-Score} \textbf{(}$\mathbf{\uparrow}$\textbf{)} & $\mathbf{\mu}$\textbf{WF} \textbf{(}$\mathbf{\uparrow}$\textbf{)} & \textbf{Dice} \textbf{(}$\mathbf{\uparrow}$\textbf{)} &  \textbf{e-measure} \textbf{(}$\mathbf{\uparrow}$\textbf{)} & \textbf{s-measure} \textbf{(}$\mathbf{\uparrow}$\textbf{)} & \textbf{MAE} \textbf{(}$\mathbf{\downarrow}$\textbf{)}\\
\hline \hline

SAMNet & 0.518 $\pm$ 0.209 & 0.431 $\pm$ 0.262 & 0.440 $\pm$ 0.256 & 0.516 $\pm$ 0.276 & 0.478 $\pm$ 0.257 & 0.418 $\pm$ 0.242\\
MSCNet & 0.730 $\pm$ 0.032 & 0.695 $\pm$ 0.009 & 0.709 $\pm$ 0.008 & 0.854 $\pm$ 0.017 & 0.775 $\pm$ 0.009 & 0.086 $\pm$ 0.023\\
MEANet & 0.715 $\pm$ 0.039 & 0.655 $\pm$ 0.034 & 0.666 $\pm$ 0.034 & 0.797 $\pm$ 0.044 & 0.733 $\pm$ 0.035 & 0.150 $\pm$ 0.047\\

\hline \hline

FLIM & 0.821 $\pm$ 0.010 & 0.660 $\pm$ 0.011 & 0.680 $\pm$ 0.010 & 0.848 $\pm$ 0.014 & 0.756 $\pm$ 0.012 & 0.057 $\pm$ 0.007 \\
FLIM\_MCA & 0.834 $\pm$ 0.017 & 0.816 $\pm$ 0.017 & 0.823 $\pm$ 0.015 & 0.940 $\pm$ 0.004 & 0.879 $\pm$ 0.008 & 0.031 $\pm$ 0.008 \\

\hline \hline

BasNet & 0.701 $\pm$ 0.118 & 0.654 $\pm$ 0.128 & 0.663 $\pm$ 0.126 & 0.786 $\pm$ 0.111 & 0.737 $\pm$ 0.109 & 0.155 $\pm$ 0.094\\
U$\mathbf{^2}$-Net & 0.687 $\pm$ 0.085 & 0.617 $\pm$ 0.089 & 0.631 $\pm$ 0.089 & 0.760 $\pm$ 0.074 & 0.716 $\pm$ 0.074 & 0.158 $\pm$ 0.062\\
\hline \hline
\end{tabular*}\label{tab:parasites_bench}
\end{table*}

The merging network's potential is also confirmed when applied to FLIM-Network saliencies. Even before CA, the merge network can improve the saliency map (e.g., 17.5\% improvement on Dice). Besides exploring each saliency map's best features, the merge network also facilitates inferences using FLIM networks. When extracting multi-level saliency maps, we need to verify which layer yields better saliency (\eg we have a higher average dice on layer 3). Merging multiple saliencies eliminates this need. Also, as inference time is a requisite, these three convolutions can be converted to simple arithmetic operations executed pixel-wise once.

\begin{figure}[!ht]
    \centering 
    \includegraphics[width=0.48\textwidth]{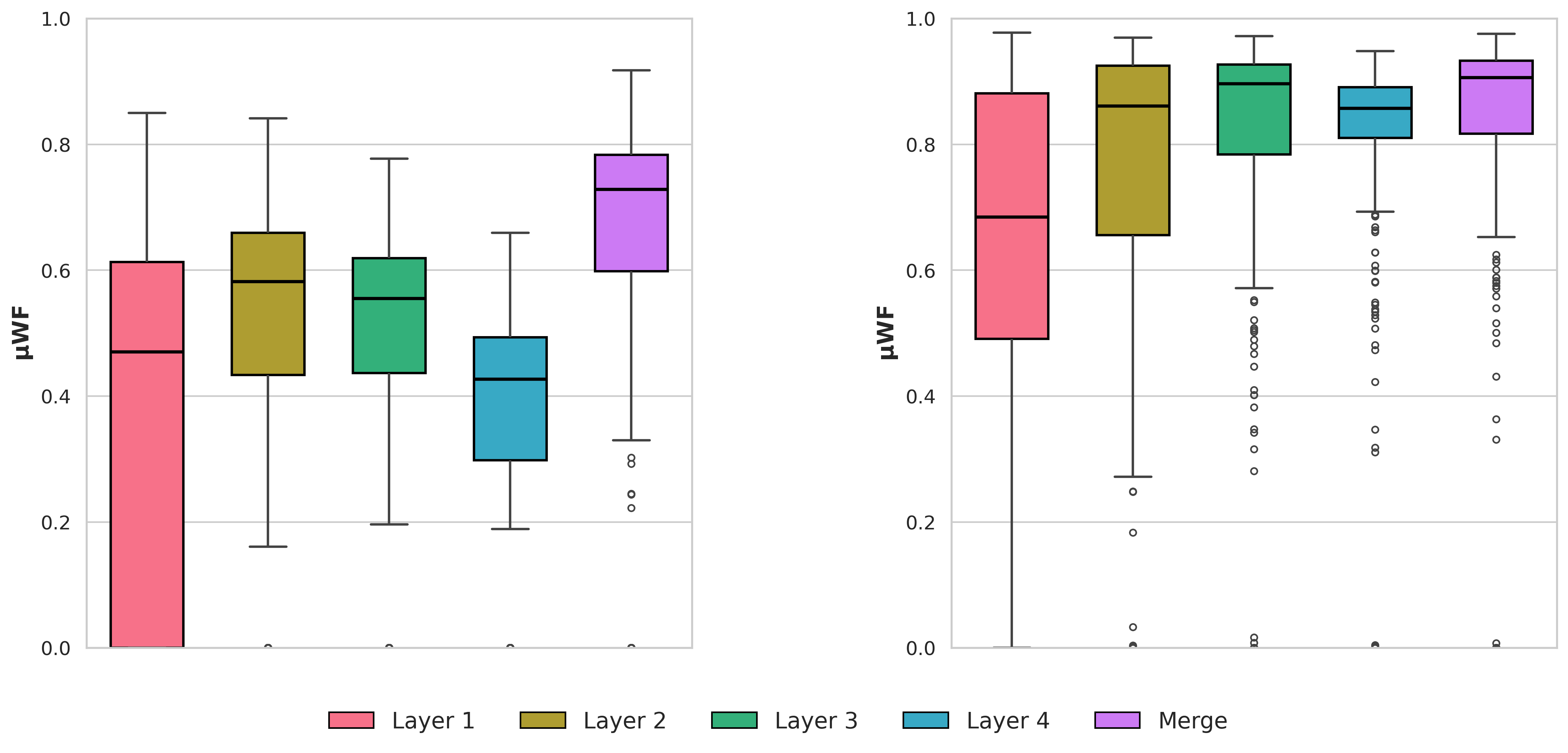}
    \caption{Metrics distribution across all layers (parasites data) on split 1. FLIM distribution on the left against FLIM\_MCA on the right.}
\label{fig:comparison_box_plot}
\end{figure}

When we analyze a split, the same conclusions arise. Figure \ref{fig:comparison_box_plot} shows metrics distribution ($\mu$WF) across each layer, and after merging, for split 1. An expressive improvement in the metric average and quartiles is observed, which ensures better stability after running CA. More interestingly, we see the FLIM key feature to extract features and, consequently, saliencies, which shows a delineation-to-detection tendency. Initial layers have more evident edges that fit ground-truth masks but may present many false positives, which results in the higher standard deviation of layer 1. As we go deeper into the network, edges get blurrier. However, false positives tend to disappear and, consequently, are better for CA initialization, as the evolution tends to correct edge regions if it has lower confidence (i.e., blurrier). This behavior is evident in image \ref{fig:grid_parasites} (\ref{fig:grid_parasites_img2}-\ref{fig:grid_parasites_img5}).

Table \ref{tab:parasites_bench} then compares merged metrics (both FLIM and FLIM\_CA) with lightweight and conventional deep SOD. We see that FLIM merged saliencies are comparable to both deep SOD methods, while FLIM\_CA outperforms all methods. Moreover, lightweight models (MSCNet and MEANet) also outperform deeper models, which shows that under constrained scenarios (few training images), the optimization over a more complex surface (more model parameters) is problematic, even with models pretrained on larger datasets (i.e., DUTS-TR or ImageNet). As previously stated, lightweights target scenarios with limited computing resources, but when compared to FLIM-Networks, models are yet larger. Take, for instance, MSCNet, with 3.6M parameters, against a FLIM network with ~400k (~11.11\%) parameters for split 1, which substantially outperforms MSCNet when applying multi-level CA. FLIM\_MCA is a post-processing operation that does not increase the parameter count.

\begin{table*}[!ht]
\caption{Metrics for BraTS 2D on test set.}
\centering
\begin{tabular*}{\textwidth}{@{}c*{6}{@{\extracolsep{\fill}}c}@{}}
\hline \hline
\textbf{Metric} & \textbf{Model} & \textbf{Layer\_1}   & \textbf{Layer\_2}     &  \textbf{Layer\_3}  &  \textbf{Merged}\\
\hline \hline

\multirow{2}{*}{\begin{tabular}[c]{@{}c@{}}\textbf{F-Score}\\\textbf{(}$\mathbf{\uparrow}$\textbf{)}\end{tabular}}
& FLIM & 0.516 $\pm$ 0.015 & 0.633 $\pm$ 0.022 & 0.673 $\pm$ 0.016 & 0.699 $\pm$ 0.034 \\
& FLIM\_MCA & 0.617 $\pm$ 0.007 & 0.718 $\pm$ 0.013 & 0.730 $\pm$ 0.015 & 0.712 $\pm$ 0.005 \\
\hline

\multirow{2}{*}{\begin{tabular}[c]{@{}c@{}}$\mathbf{\mu}$\textbf{WF}\\\textbf{(}$\mathbf{\uparrow}$\textbf{)}\end{tabular}}
& FLIM & 0.577 $\pm$ 0.021 & 0.675 $\pm$ 0.020 & 0.688 $\pm$ 0.011 & 0.697 $\pm$ 0.013 \\
& FLIM\_MCA & 0.642 $\pm$ 0.012 & 0.698 $\pm$ 0.012 & 0.656 $\pm$ 0.025 & 0.710 $\pm$ 0.013 \\
\hline

\multirow{2}{*}{\begin{tabular}[c]{@{}c@{}}\textbf{Dice}\\\textbf{(}$\mathbf{\uparrow}$\textbf{)}\end{tabular}}
& FLIM & 0.574 $\pm$ 0.021 & 0.675 $\pm$ 0.018 & 0.689 $\pm$ 0.011 & 0.700 $\pm$ 0.015 \\
& FLIM\_MCA & 0.645 $\pm$ 0.012 & 0.701 $\pm$ 0.011 & 0.664 $\pm$ 0.024 & 0.713 $\pm$ 0.011 \\
\hline

\multirow{2}{*}{\begin{tabular}[c]{@{}c@{}}\textbf{e-measure}\\\textbf{(}$\mathbf{\uparrow}$\textbf{)}\end{tabular}}
& FLIM & 0.707 $\pm$ 0.035 & 0.822 $\pm$ 0.018 & 0.848 $\pm$ 0.015 & 0.841 $\pm$ 0.018 \\
& FLIM\_MCA & 0.822 $\pm$ 0.001 & 0.870 $\pm$ 0.005 & 0.833 $\pm$ 0.020 & 0.877 $\pm$ 0.007 \\
\hline

\multirow{2}{*}{\begin{tabular}[c]{@{}c@{}}\textbf{s-measure}\\\textbf{(}$\mathbf{\uparrow}$\textbf{)}\end{tabular}}
& FLIM & 0.624 $\pm$ 0.033 & 0.726 $\pm$ 0.013 & 0.756 $\pm$ 0.012 & 0.771 $\pm$ 0.017 \\
& FLIM\_MCA & 0.754 $\pm$ 0.006 & 0.787 $\pm$ 0.008 & 0.757 $\pm$ 0.017 & 0.799 $\pm$ 0.009 \\
\hline

\multirow{2}{*}{\begin{tabular}[c]{@{}c@{}}\textbf{MAE}\\\textbf{(}$\mathbf{\downarrow}$\textbf{)}\end{tabular}}
& FLIM & 0.166 $\pm$ 0.057 & 0.082 $\pm$ 0.012 & 0.069 $\pm$ 0.016 & 0.075 $\pm$ 0.021 \\
& FLIM\_MCA & 0.044 $\pm$ 0.001 & 0.027 $\pm$ 0.002 & 0.034 $\pm$ 0.004 & 0.032 $\pm$ 0.001 \\
\hline

\hline \hline
\end{tabular*}\label{tab:brain_all}
\end{table*}

\begin{table*}[!ht]
\caption{Benchmark on BraTS 2D test set (FLIM Metrics stands for merged model).}
\centering
\begin{tabular*}{\textwidth}{@{}c*{6}{@{\extracolsep{\fill}}c}@{}}
\hline \hline
\textbf{Model} & \textbf{F-Score} \textbf{(}$\mathbf{\uparrow}$\textbf{)} & $\mathbf{\mu}$\textbf{WF} \textbf{(}$\mathbf{\uparrow}$\textbf{)} & \textbf{Dice} \textbf{(}$\mathbf{\uparrow}$\textbf{)} &  \textbf{e-measure} \textbf{(}$\mathbf{\uparrow}$\textbf{)} & \textbf{s-measure} \textbf{(}$\mathbf{\uparrow}$\textbf{)} & \textbf{MAE} \textbf{(}$\mathbf{\downarrow}$\textbf{)}\\
\hline \hline

SAMNet & 0.000 $\pm$ 0.000 & 0.000 $\pm$ 0.000 & 0.000 $\pm$ 0.000 & 0.000 $\pm$ 0.000 & 0.000 $\pm$ 0.000 & 1.000 $\pm$ 0.000\\
MSCNet & 0.431 $\pm$ 0.012 & 0.234 $\pm$ 0.003 & 0.269 $\pm$ 0.001 & 0.459 $\pm$ 0.028 & 0.509 $\pm$ 0.003 & 0.158 $\pm$ 0.017\\
MEANet & 0.118 $\pm$ 0.020 & 0.116 $\pm$ 0.013 & 0.123 $\pm$ 0.014 & 0.680 $\pm$ 0.054 & 0.446 $\pm$ 0.036 & 0.149 $\pm$ 0.078\\

\hline \hline

FLIM &  0.699 $\pm$ 0.034 & 0.697 $\pm$ 0.013 & 0.700 $\pm$ 0.015 & 0.841 $\pm$ 0.018 & 0.771 $\pm$ 0.017 & 0.075 $\pm$ 0.021 \\
FLIM\_MCA & 0.712 $\pm$ 0.005 & 0.710 $\pm$ 0.013 & 0.713 $\pm$ 0.011 & 0.877 $\pm$ 0.007 & 0.799 $\pm$ 0.009 & 0.032 $\pm$ 0.001\\

\hline \hline

BasNet & 0.775 $\pm$ 0.022 & 0.751 $\pm$ 0.020 & 0.756 $\pm$ 0.019 & 0.880 $\pm$ 0.028 & 0.830 $\pm$ 0.011 & 0.024 $\pm$ 0.007\\
U$\mathbf{^2}$-Net & 0.728 $\pm$ 0.026 & 0.684 $\pm$ 0.017 & 0.692 $\pm$ 0.015 & 0.834 $\pm$ 0.033 & 0.785 $\pm$ 0.012 & 0.037 $\pm$ 0.012\\
\hline \hline
\end{tabular*}\label{tab:brain_bench}
\end{table*}

Moving our analysis to the problem of detecting saliency, which stands for whole tumors on the BraTS 2D dataset, the same steps were adopted to design the FLIM encoder. Despite holding grayscale images, the BraTS 2D dataset is challenging, given data heterogeneity (as the original 3D dataset is multi-institutional) and the necessity for domain adaptation during fine-tuning (pretrained models were developed on RGB images).

\begin{figure}[!ht]
    \centering 
    \includegraphics[width=0.48\textwidth]{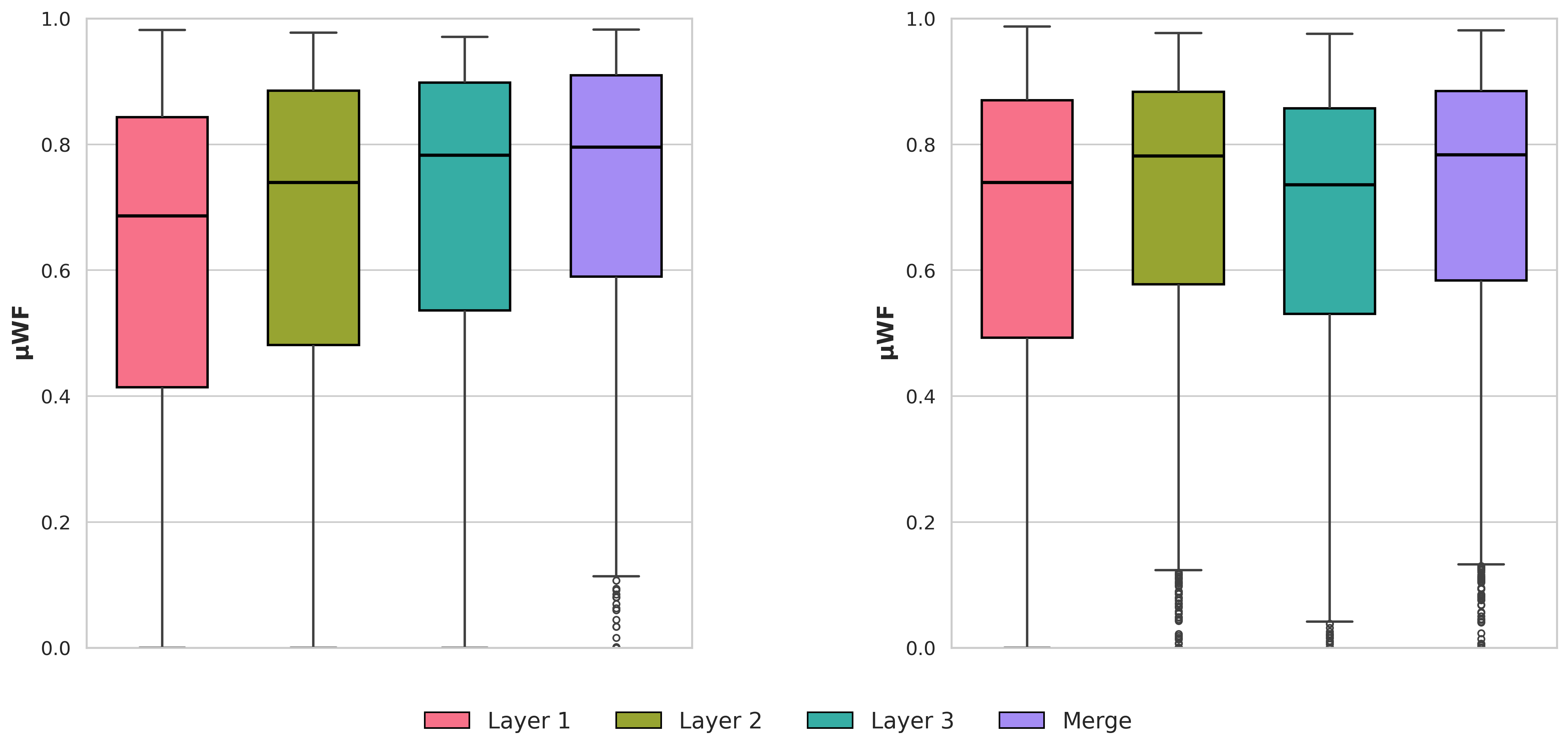}
    \caption{Metrics distribution across all layers (BraTS 2D data) on split 2. FLIM distribution on the left against FLIM\_MCA on the right.}
\label{fig:comparison_box_plot_brain}
\end{figure}

Contrary to parasite experiments, except for F-Score and MAE, Table \ref{tab:brain_all} shows improvements on BraTS 2D only on the first and second layers. Nevertheless, gains in metrics are substantial: F-Score (by 5.6-10.1\%), $\mu$WF (by 2.2-6.5\%), Dice coefficient (by 2.5-7.1\%), e-measure (by 4.8-11.5\%), and s-measure (by 6.1-13\%), and on MAE we see and error reduction of 0.035 to 0.122. An analysis of a split's distribution (Figure \ref{fig:comparison_box_plot_brain}) shows that the method saturates when it reaches layer 3. However, for all other layers, there is an improvement in both average metric value and stability (standard deviation), while some outliers remain (dots on the image). Still, we see a higher standard deviation across all layers for every split, which we believe is strongly related to data heterogeneity; hence, selecting more representative images for the FLIM encoder design is important.

Furthermore, as we see saturation on layer 3, experimenting with minor modifications to architecture design is also important. Examples are adding stride on layers 2 and 3 (pooling) and experimenting with average pooling. As overconfidence in edge regions causes the CA to leak the background into the foreground, blurrier edges would be of great importance and have the potential to overcome saturation. Also, a significant difference from our previous work is how we threshold the object probability map \citep{crispim_sibgrapi_2024}. Instead of a fixed threshold, we employed a histogram-based threshold, significantly improving our results. We omitted those results for conciseness but experimented with fixed thresholds, the Otsu threshold computed using the brain mask, and different histogram-based thresholds. In summary, all approaches generated saliencies worse than FLIM. It emphasizes the key role of thresholding for tumor salient object detection, and there is room for improvement.

\begin{figure}[!ht]
    \centering 
    \includegraphics[width=0.24\textwidth]{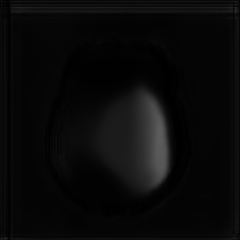}
    \caption{Sample of SAMNet's predictions.}
\label{fig:problematic_sam_net}
\end{figure}

\begin{figure*}[!th]
    \centering
    % Row 1
    \subfloat[Input image]{\includegraphics[width=0.18\textwidth]{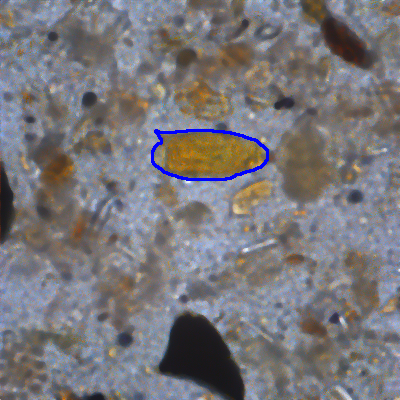}\label{fig:grid_parasites_img1}}
    \hfill
    \subfloat[FLIM - Layer 1]{\includegraphics[width=0.18\textwidth]{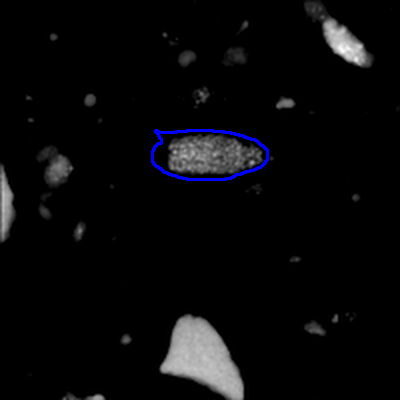}\label{fig:grid_parasites_img2}}
    \hfill
    \subfloat[FLIM - Layer 2]{\includegraphics[width=0.18\textwidth]{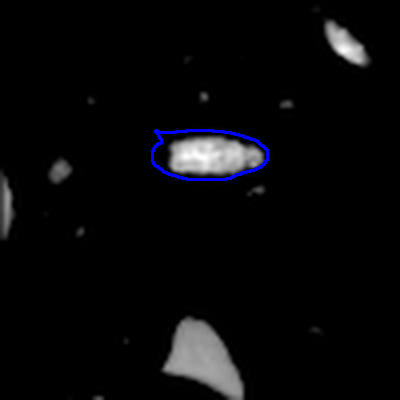}\label{fig:grid_parasites_img3}}
    \hfill
    \subfloat[FLIM - Layer 3]{\includegraphics[width=0.18\textwidth]{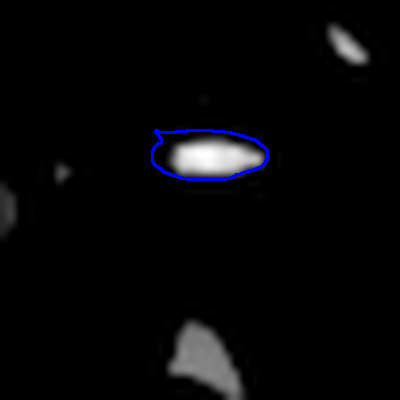}\label{fig:grid_parasites_img4}}
    \hfill
    \subfloat[FLIM - Layer 4]{\includegraphics[width=0.18\textwidth]{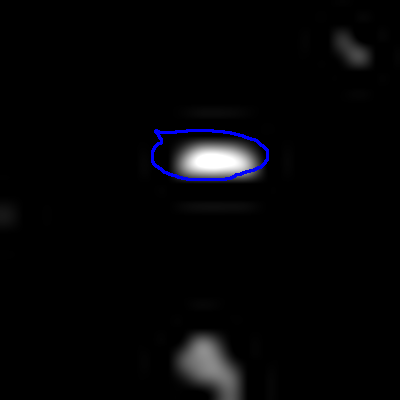}\label{fig:grid_parasites_img5}}
    
    % Row 2
    \subfloat[Merged Saliency]{\includegraphics[width=0.18\textwidth]{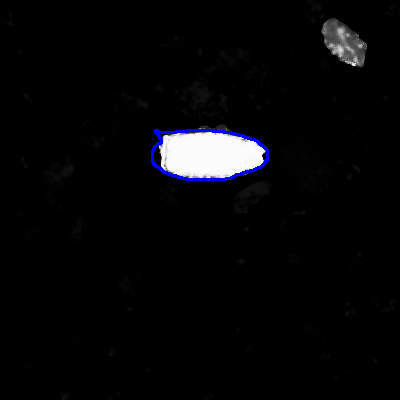}\label{fig:grid_parasites_img6}}
    \hfill
    \subfloat[FLIM\_MCA - Layer 1]{\includegraphics[width=0.18\textwidth]{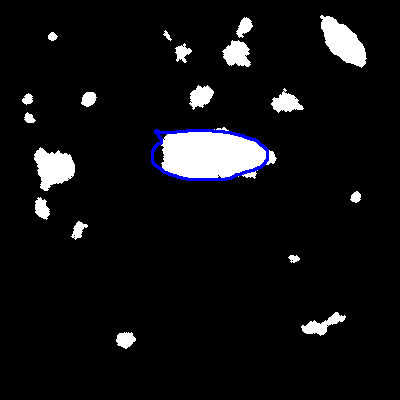}\label{fig:grid_parasites_img7}}
    \hfill
    \subfloat[FLIM\_MCA - Layer 2]{\includegraphics[width=0.18\textwidth]{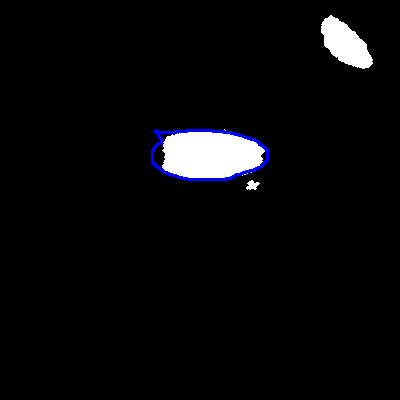}\label{fig:grid_parasites_img8}}
    \hfill
    \subfloat[FLIM\_MCA - Layer 3]{\includegraphics[width=0.18\textwidth]{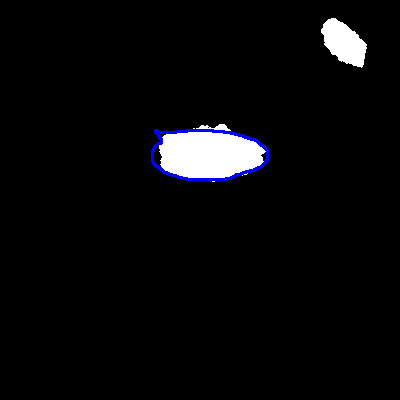}\label{fig:grid_parasites_img9}}
    \hfill
    \subfloat[FLIM\_MCA - Layer 4]{\includegraphics[width=0.18\textwidth]{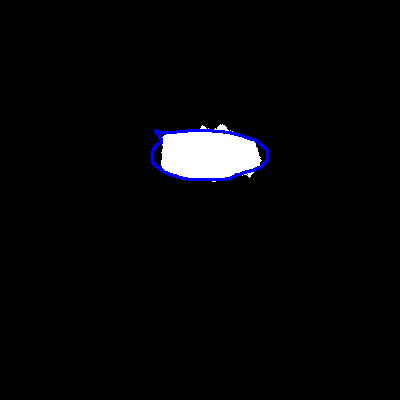}\label{fig:grid_parasites_img10}}
    
    % Row 3
    \subfloat[Input image]{\includegraphics[width=0.18\textwidth]{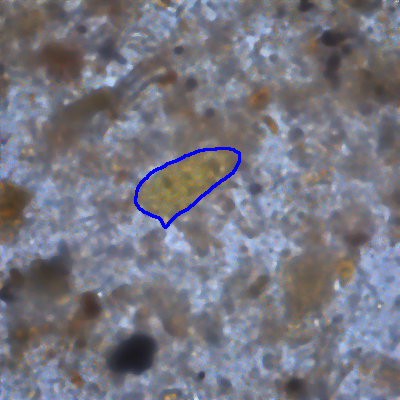}\label{fig:grid_parasites_img11}}
    \hfill
    \subfloat[FLIM - Layer 1]{\includegraphics[width=0.18\textwidth]{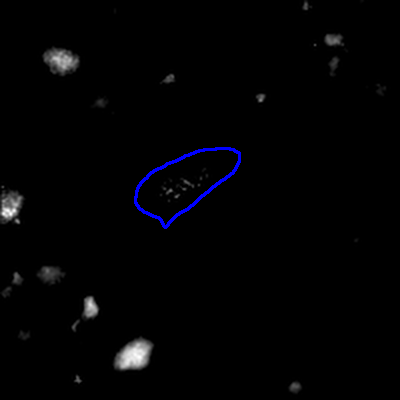}\label{fig:grid_parasites_img12}}
    \hfill
    \subfloat[FLIM - Layer 2]{\includegraphics[width=0.18\textwidth]{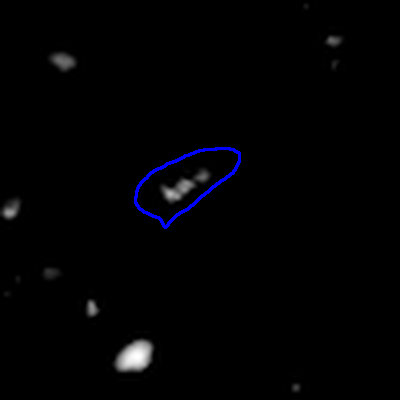}\label{fig:grid_parasites_img13}}
    \hfill
    \subfloat[FLIM - Layer 3]{\includegraphics[width=0.18\textwidth]{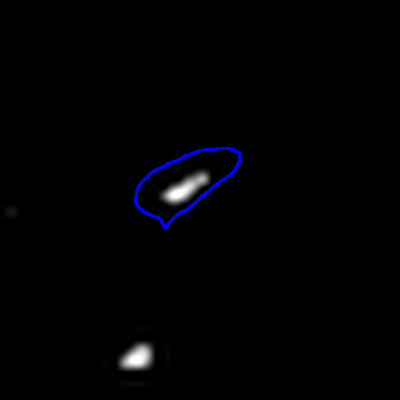}\label{fig:grid_parasites_img14}}
    \hfill
    \subfloat[FLIM - Layer 4]{\includegraphics[width=0.18\textwidth]{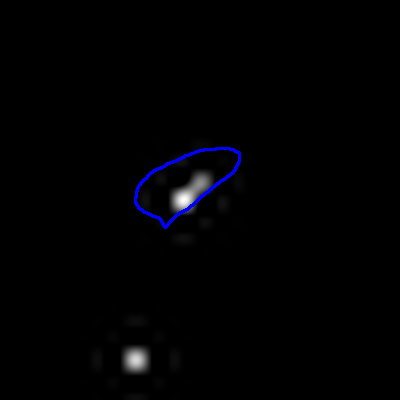}\label{fig:grid_parasites_img15}}
    
    % Row 4
    \subfloat[Merged Saliency]{\includegraphics[width=0.18\textwidth]{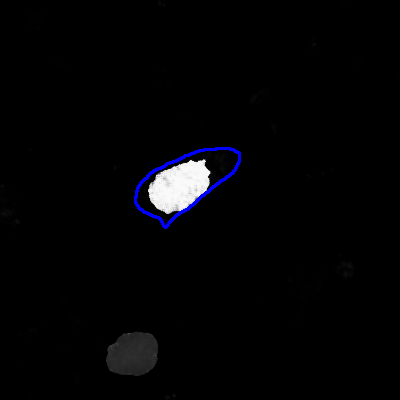}\label{fig:grid_parasites_img16}}
    \hfill
    \subfloat[FLIM\_MCA - Layer 1]{\includegraphics[width=0.18\textwidth]{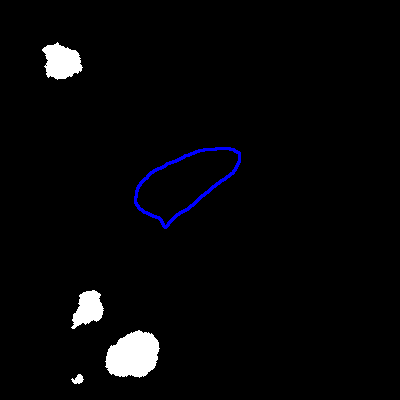}\label{fig:grid_parasites_img17}}
    \hfill
    \subfloat[FLIM\_MCA - Layer 2]{\includegraphics[width=0.18\textwidth]{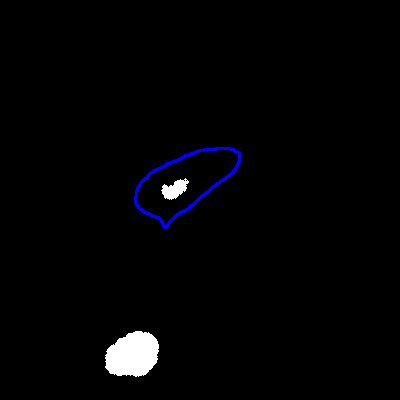}\label{fig:grid_parasites_img18}}
    \hfill
    \subfloat[FLIM\_MCA - Layer 3]{\includegraphics[width=0.18\textwidth]{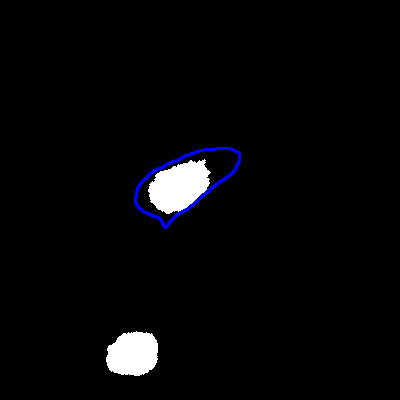}\label{fig:grid_parasites_img19}}
    \hfill
    \subfloat[FLIM\_MCA - Layer 4]{\includegraphics[width=0.18\textwidth]{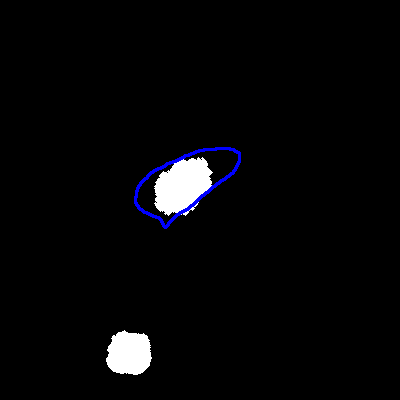}\label{fig:grid_parasites_img20}}
    
    \caption{Multi-level CA on parasite data. The first two rows show images with superior performance, while the bottom rows show problematic ones. The blue line represents ground-truth mask edges.}
    \label{fig:grid_parasites}
\end{figure*}

\begin{figure*}[!th]
    \centering
    % Row 1
    \subfloat[Input image]{\includegraphics[width=0.23\textwidth]{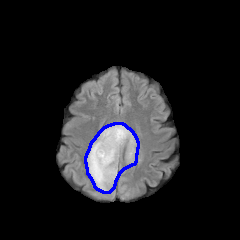}\label{fig:grid_brain_img1}}
    \hfill
    \subfloat[FLIM - Layer 1]{\includegraphics[width=0.23\textwidth]{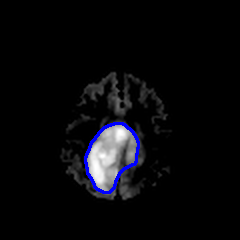}\label{fig:grid_brain_img2}}
    \hfill
    \subfloat[FLIM - Layer 2]{\includegraphics[width=0.23\textwidth]{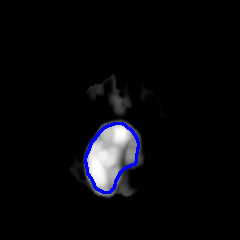}\label{fig:grid_brain_img3}}
    \hfill
    \subfloat[FLIM - Layer 3]{\includegraphics[width=0.23\textwidth]{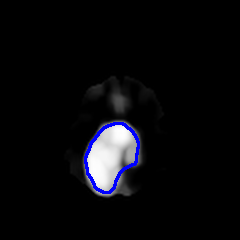}\label{fig:grid_brain_img4}}
    
    % Row 2
    \subfloat[Merged Saliency]{\includegraphics[width=0.23\textwidth]{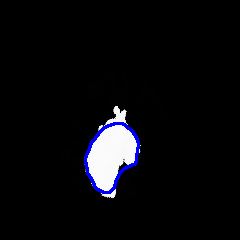}\label{fig:grid_brain_img5}}
    \hfill
    \subfloat[FLIM\_MCA - Layer 1]{\includegraphics[width=0.23\textwidth]{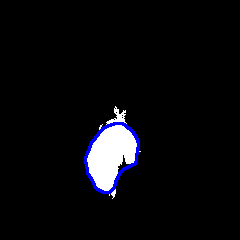}\label{fig:grid_brain_img6}}
    \hfill
    \subfloat[FLIM\_MCA - Layer 2]{\includegraphics[width=0.23\textwidth]{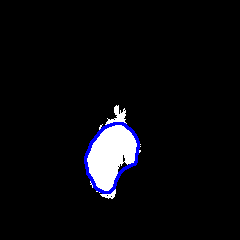}\label{fig:grid_brain_img7}}
    \hfill
    \subfloat[FLIM\_MCA - Layer 3]{\includegraphics[width=0.23\textwidth]{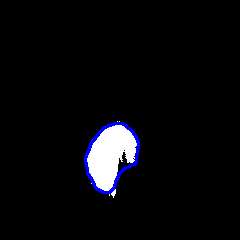}\label{fig:grid_brain_img8}}
    
    % Row 3
    \subfloat[Input image]{\includegraphics[width=0.23\textwidth]{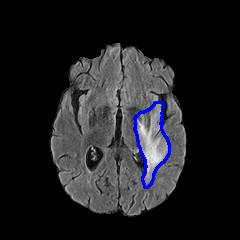}\label{fig:grid_brain_img9}}
    \hfill
    \subfloat[FLIM - Layer 1]{\includegraphics[width=0.23\textwidth]{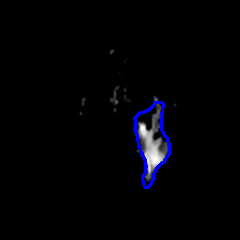}\label{fig:grid_brain_img10}}
    \hfill
    \subfloat[FLIM - Layer 2]{\includegraphics[width=0.23\textwidth]{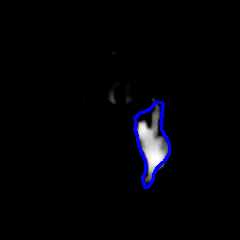}\label{fig:grid_brain_img11}}
    \hfill
    \subfloat[FLIM - Layer 3]{\includegraphics[width=0.23\textwidth]{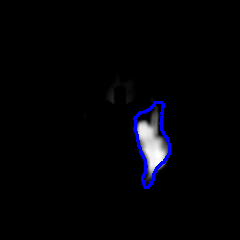}\label{fig:grid_brain_img12}}
    
    % Row 4
    \subfloat[Merged Saliency]{\includegraphics[width=0.23\textwidth]{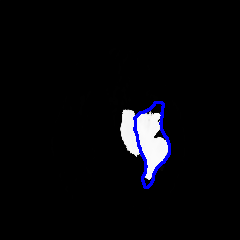}\label{fig:grid_brain_img13}}
    \hfill
    \subfloat[FLIM\_MCA - Layer 1]{\includegraphics[width=0.23\textwidth]{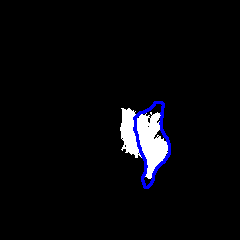}\label{fig:grid_brain_img14}}
    \hfill
    \subfloat[FLIM\_MCA - Layer 2]{\includegraphics[width=0.23\textwidth]{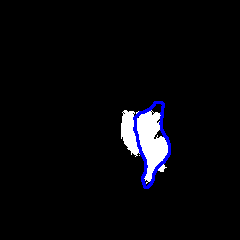}\label{fig:grid_brain_img15}}
    \hfill
    \subfloat[FLIM\_MCA - Layer 3]{\includegraphics[width=0.23\textwidth]{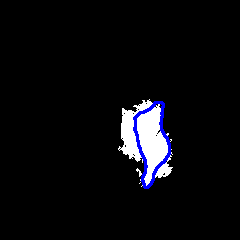}\label{fig:grid_brain_img16}}
    
    \caption{Multi-level CA on BraTS 2D data. The first two rows show images with superior performance, while the bottom rows show problematic ones. The blue line represents ground-truth mask edges.}
    \label{fig:grid_brain}
\end{figure*}

When we merge the multi-level saliencies, the same improvements observed on parasites stand out (see also Figure \ref{fig:comparison_box_plot_brain}. However, on FLIM saliencies, for e-measure and MAE, merge statistics are closer to the best layer but do not surpass it. Merging FLIM\_MCA saliencies also yields better metrics, except for F-Score and MAE, which are comparable to the best layer. Both results show that in the worst scenario, merging yields comparable results. Merging results on both datasets lay down the floor for further experiments with merging, as it shows that taking advantage of all layers' saliencies is possible and could improve overall results. Experiments with more complex networks, taking the proper care to avoid overfitting, have the potential to further improvements. Moreover, using the probability map instead of the binarized saliency could eliminate the necessity to compute thresholds. We tried the same merge networks using the probability map but without success.

Comparing merged results with deep SOD literature shows a different behavior than the parasites' benchmark. First, lightweight networks fail fine-tuning to a different domain --- pre-trained weights were learned on RGB images --- when few training images are available. All networks fail to learn, even with our efforts to experiment with different learning rates to reach the best performance. SAMNet could not predict saliency maps, which detect tumors correctly, and saliency maps had all values below the threshold for metric computation. Thus, SAMNet has the worst metrics possible.

Distinctly from lightweight methods, deeper networks learned a different domain (gray-scale images), in which BasNet and U$\mathbf{^2}$-Net generate better saliency maps than all lightweight models. BasNet has the best results. Nevertheless, FLIM and FLIM\_MCA showed competitive results compared to BasNet, significantly outperforming U$\mathbf{^2}$-Net on most metrics.

From a qualitative perspective, Figures \ref{fig:grid_parasites} and \ref{fig:grid_brain} summarize our observations from a detailed analysis of the method outputs. Both images present the FLIM output (CA initialization), the CA output, and the merged saliency. We also provide two examples: one demonstrating a successful case and the other illustrating a problematic one. As a blue line, we see ground-truth edges.

Figure \ref{fig:grid_parasites} (\ref{fig:grid_parasites_img1}-\ref{fig:grid_parasites_img10}) shows visualize what we have already discussed so far. First, on layer 1, we see much noise (false positives) while the parasite's egg itself is darker than some false positives. Deeper saliency values on the egg surpass false positive values, while false positives disappear. Finally, we see the blurrier characteristic on the last layer and only two false-positive components. Despite the noise of initialization, we see that both the initialization and evolution procedures are robust and can eliminate most false positives (dark debris at the bottom of the image). On the last layer, CA eliminates all false positives. The merged saliency has a single false positive component, and the only remaining (which appears on layers 1 to 3) has a lower saliency value and is eliminated in the thresholding step. Supporting our discussion on the method's robustness against false positives, Figures \ref{fig:grid_parasites_img11} to \ref{fig:grid_parasites_img20} show how initialization and evolution eliminate most of the wrong saliencies. We clearly see that the darker component will be thresholded out on merged saliency. However, it shows a limitation of our method on evolution when lower saliency values are employed for initialization or when only a small portion of the parasite egg was salient on FLIM saliencies. The CA cannot evolve correctly to fit most parasites' eggs in such cases. A possible approach to overcome this problem is extracting components' statistics (e.g., mean LAB value on salient component) and using this description to propagate cells' strengths better.

For brain tumor examples (Figures \ref{fig:grid_brain_img1} to \ref{fig:grid_brain_img8}), we see that despite false positives, our method correctly fits the whole tumor, eliminating false positive values. On the brain, false positives consistently have lower values than on tumors. Furthermore, on FLIM saliencies for layer 2 and layer 3, we can spot higher saliency values outside the tumor. Nevertheless, as we initialize background strength on the brain border and evolve it into the brain, we correctly eliminate those regions. The second case shows a more problematic example, which is frequent on training data. If we analyze the input image, we can easily see that the tumor ground truth encompasses a hyper-intense region but a dark one, too. As it has intensity values similar to brain tumor tissue, FLIM saliencies also detect the dark region as a tumor, and evolution happens, leaking into the background. The leaking is more intense on deeper layers, as more regions are initialized with higher values, which justifies the FLIM\_MCA degradation on the third layer.

Leaking may occur when the CA propagates foreground strength into healthy brain tissue or background regions on parasites due to incorrect initialization. More typically, when using deeper network layers (like layer 3), their saliency maps are blurrier and cover larger areas than those from earlier layers. This saliency causes healthy brain regions (hypo-intense regions in Figure \ref{fig:grid_brain_img9}) to be incorrectly initialized as foreground (tumor). During evolution, these misclassified regions act as ``seeds'' that spread and ``conquer'' neighboring tissues with similar characteristics, resulting in false positives extending beyond the tumor boundaries. A possible approach for better robustness would be experimenting with different pooling (average or minimum), which could provide a better initialization. It is important to note that the ground-truth tumor masks annotated by specialists also include these hypo-intense regions (blue line on Figures \ref{fig:grid_parasites} and \ref{fig:grid_brain}).

From the results discussed, a conclusion arises. With a fraction of parameters, FLIM\_MCA outperforms deep SOD methods for parasites and reaches competitive performances under a more heterogeneous problem (BraTS 2D). Nevertheless, there is plenty of room for improvements in every phase of the proposed method, from image selection to decode, initialization, evolution, and merging. We also note that the implemented solution employs Open MP, where the whole pipeline (CA initialization, evolution, and merging) takes around 3s. There is significant potential to improve efficiency when working with constrained computing resources.

\section{Conclusion}
\label{sec:6}

We expanded previous work on improving the CA initialization problem by exploring knowledge distilled at different hierarchical levels of FLIM networks. CA methods have shown competitive performance, particularly in medical problems where abundant training and annotated data are challenging to obtain. While CA typically requires user interaction for every image, our approach incorporates user knowledge into a FLIM network that initializes the CA automatically. By initializing a separate CA for each convolutional layer of the FLIM Network, we create a multi-level CA that explores a broader initialization space, consistently improving saliency at each level (up to 43.4\%).

The multi-level approach generates multiple complementary saliency maps that form a CA ensemble, leveraging each encoder level's unique characteristics --- from sharp-edged features with false positives to blurrier edges with fewer false positives. Our experiments demonstrate that these saliency maps can be effectively combined using a straightforward merging network (just three convolutional filters) trained on only 3-4 images. The merging network eliminates the need to determine which network layer yields the best saliency while maintaining high performance.

Our method showed consistent cross-domain improvements when benchmarked across both public (BraTS) and private (parasite eggs detection) datasets. For parasite detection, multi-level CA outperformed all compared models (lightweight and deeper architectures). For brain tumor detection, we outperformed all lightweight models and U²-Net while achieving comparable results to BasNet - despite using only a fraction of the parameters (400k versus 3.6M in lightweight models) and requiring only 3-4 weakly-annotated training images compared to the thousands or millions used by benchmarked approaches.

As a limitation, we see the separate evolution of foreground and background CA as an additional computational overhead that could be optimized. The method also shows some performance variability across datasets, with more substantial results on parasite detection than on more heterogeneous brain tumor data.
% Where - Try substitute for in which!
For future work, we plan to evolve a single CA instead of separate foreground and background automata, which would reduce computational overhead and simplify rule design. This approach could enable more complex cellular interactions and facilitate 3D multi-label CA development for semantic tumor segmentation - correctly evolving a CA to simultaneously segment the whole tumor, enhanced tumor, and edematous tissue, which could be tried under a clinical environment. The simplicity of our merging network offers substantial room for investigating optimal methods to combine multi-level salience/segmentation maps. Additionally, exploring methods for smoothing border surfaces could further improve edge regions \citep{improved_ca, grow_cut}.

From an implementation perspective, our current solution leverages CPU parallelization (OpenMP); significant speed improvements are possible through GPU optimization targeting low-cost hardware suitable for real-world applications like parasitology laboratories in developing regions.

\section*{Acknowledgment}
We thank the Eldorado Institute. Moreover, this work was financially supported by the Conselho Nacional de Desenvolvimento Cient{\'i}fico e Tecnol{\'o}gico -- CNPq --  (407242/2021-0, 306573/2022-9, 442950/2023-3, 304711/2023-3),  the Funda\c{c}{\~a}o de Amparo a Pesquisa do Estado de Minas Gerais -- FAPEMIG --  (APQ-01079-23 and APQ-05058-23), the Funda\c{c}{\~a}o de Amparo a Pesquisa do Estado de S{\~a}o Paulo -- FAPESP -- (2023/14427-8 and  2013/07375-0) and the Coordena{\c c}{\~a}o de Aperfei{\c c}oamento de Pessoal de N{\'i}vel Superior -- CAPES -- (STIC-AMSUD 88887.878869/2023-00).

\end{document}